%% file: main.tex
\documentclass[10pt,twocolumn,letterpaper]{article}

\usepackage[pagenumbers]{cvpr}
\input{preamble}

\definecolor{cvprblue}{rgb}{0.21,0.49,0.74}
\usepackage[breaklinks,colorlinks,allcolors=cvprblue]{hyperref}

\title{DFM-WAM: Discrete Flow Matching World--Action Models from Pretraining to Post-Training}

\author{
  \textbf{Yitian Liu$^{1,3*}$,\ Yuhao Wu$^{2,3*}$,\ Weijie Shen$^{1}$\thanks{Equal contribution. $^{\dagger}$Corresponding author. $^{\ddagger}$Project leader.}}\\
  \textbf{Zhixuan Liang$^{4}$,\ Congsheng Xu$^{1}$,\ Haotian Liang$^{5}$,\ Wenjie Xu$^{3}$,\ Sheng Zhang$^{3}$}\\
  \textbf{Zhongshan Liu$^{3}$,\ Yinan Mao$^{3}$,\ Lei Xu$^{1}$,\ Xinping Guan$^{1}$,\ Ru Ying$^{3\ddagger}$,\ Ran Zheng$^{3}$}\\
  \textbf{Xiaokang Yang$^{1}$,\ Wenbo Ding$^{2}$,\ Yao Mu$^{1\dagger}$}\\
  $^1$Shanghai Jiao Tong University\\
  $^2$Tsinghua Shenzhen International Graduate School, Tsinghua University\\
  $^3$Baidu AI Cloud \quad $^4$Princeton University \quad $^5$University of Hong Kong\vspace{-6pt}
}

\begin{document}
\maketitle
\renewcommand{\endabstract}{\par}
\input{sec/0_abstract}
\input{sec/1_intro}

\input{sec/2_related_work}
\input{sec/3_method}
\input{sec/4_experiments}
\input{sec/5_conclusion}

\subsubsection*{Acknowledgments}
We would like to express our sincere gratitude to the Baidu AI Cloud Baige Team for their exceptional technical support and for providing access to the state-of-the-art Baidu AIHC platform. We specifically appreciate the platform's powerful capabilities in delivering efficient training acceleration and enabling ultra-low-latency inference, which were instrumental in optimizing our system's performance during evaluation. The advanced system optimizations and robust distributed infrastructure provided by this team were crucial in accelerating our experiments and validating the scalability of our proposed methods.

{
    \small
    \ifdefined\XeTeXversion
      \catcode"2013=\active
      \begingroup\lccode`\~="2013 \lowercase{\endgroup\def~}{\textendash}
    \fi
    \bibliographystyle{ieeenat_fullname}
    \bibliography{main}
}

\clearpage
\appendix
\twocolumn[
  \centering
  \Large
  \textbf{Appendix}\par
  \vspace{1.0em}
]
\setcounter{table}{0}
\setcounter{figure}{0}
\setcounter{equation}{0}
\renewcommand{\thetable}{S\arabic{table}}
\renewcommand{\thefigure}{S\arabic{figure}}
\renewcommand{\theequation}{S\arabic{equation}}
\renewcommand{\theHtable}{S\arabic{table}}
\renewcommand{\theHfigure}{S\arabic{figure}}
\renewcommand{\theHequation}{S\arabic{equation}}
\renewcommand{\topfraction}{0.9}
\renewcommand{\dbltopfraction}{0.9}
\renewcommand{\bottomfraction}{0.8}
\renewcommand{\textfraction}{0.07}
\renewcommand{\floatpagefraction}{0.75}
\renewcommand{\dblfloatpagefraction}{0.75}
\input{sec/supp_model}
\input{sec/supp_pretraining}
\input{sec/supp_rl}
\input{sec/supp_simulation}
\input{sec/supp_real_world}
\input{sec/supp_results}
\input{sec/supp_baselines}

\end{document}

%% file: preamble.tex
\usepackage[T1]{fontenc}

\newcommand{\method}{DFM-WAM\xspace}

%% file: sec/0_abstract.tex
\begin{abstract}
World-Action Models (WAMs) empower robot manipulation by using training-time future visual supervision to learn action-relevant dynamics, yet prevailing WAMs rely on continuous formulations for both their video backbones and action heads. However, as discrete video generators using iterative global refinement now rival continuous diffusion in spatiotemporal modeling, a natural question arises: \textbf{can a pretrained discrete video backbone with discrete future-frame supervision equally internalize physical dynamics for control?} Meanwhile, for action generation, discrete representations avoid the temporal over-smoothing of continuous policies, enhance state reactivity, and naturally align with reinforcement learning (RL). To explore this unified discrete paradigm, we present \textbf{DFM-WAM}, a compact 2.3B model pairing a 1.7B Discrete Flow Matching (DFM) Video Expert with a 0.6B DFM Action Expert under joint cross-entropy training. We find that generic discrete video priors effectively transfer to visuomotor control, achieving \textbf{95.35\%} on LIBERO without embodied pretraining, while scaling embodied pretraining boosts average success across clean and randomized scenes on the 50-task RoboTwin 2.0 benchmark to \textbf{91.47\%} (Full) and \textbf{54.74\%} (Clean-to-Random). Furthermore, we derive exact path-level importance sampling ratios for multi-step DFM denoising via transition-kernel cancellation, enabling online RL that elevates LIBERO success to \textbf{99.7\%} and RoboTwin eight-task success from \textbf{83.25\%} to \textbf{92.97\%}. Finally, we combine dense visual-change consistency rewards with sparse task rewards, enabling joint optimization of both Video and Action experts through a single PPO objective and pushing RoboTwin eight-task success to \textbf{93.75\%}.
\end{abstract}

%% file: sec/1_intro.tex
\section{Introduction}

Robotic manipulation requires anticipating how physical actions transform the world. While Vision-Language-Action (VLA) models map current observations to control~\cite{pi0,openvla}, World-Action Models (WAMs) internalize physical dynamics by using training-time future visual supervision to learn task- and action-relevant representations~\cite{cosmospolicy,lingbotva,fastwam,imagewam}. Yet, prevailing WAMs rely on continuous formulations for both their video backbones and action heads~\cite{cosmospolicy,lingbotva,fastwam,motus}. On the visual side, whereas early autoregressive or masked discrete video models suffered from non-refinable token generation~\cite{worldvla,ursa}, discrete video generators using iterative global refinement over metric probability paths now rival continuous diffusion in spatiotemporal modeling~\cite{ursa}. This raises a natural question: \textit{can a pretrained discrete video backbone with discrete future-frame supervision equally internalize physical dynamics for control?} Meanwhile, for action generation, empirical evidence shows that discrete representations avoid the temporal over-smoothing of continuous policies, enhance state reactivity~\cite{yi2026discrete}, and naturally align with reinforcement learning (RL)~\cite{dvlarl}. Furthermore, while single-pass and autoregressive discrete policies suffer from cross-dimension factorization errors or decoding latency~\cite{openvla,worldvla,yi2026discrete} and masked diffusion VLAs~\cite{mmact,ddvla,mmadavla} face irreversible token commitment~\cite{dfmvla}, Discrete Flow Matching (DFM)~\cite{dfm} mitigates these bottlenecks via iterative full-sequence refinement~\cite{dfmvla,yi2026discrete}---yet existing DFM policies only denoise actions over static observations without world modeling.

To explore this unified discrete paradigm, we present \textbf{DFM-WAM}, a compact 2.3B model pairing a 1.7B DFM Video Expert~\cite{ursa} with a 0.6B DFM Action Expert over Finite Scalar Quantization (FSQ)~\cite{fsq} visual lattices and quantized action bins. Under joint cross-entropy training along modality-specific metric paths, DFM-WAM supervises key future-frame visual tokens~\cite{imagewam} in parallel with action tokens~\cite{fastwam}. A directed attention mask allows future-frame supervision to shape the visual backbone during training while masking future tokens from the action branch, enabling fast test-time control without generating future frames. Decoupling visual dynamics and action execution into jointly attended DFM experts also supports scalable pretraining across action-free human egocentric videos~\cite{egoverse,egodex} and multi-embodiment robot data.

Beyond supervised learning, DFM-WAM unlocks tractable online RL for policy calibration and joint expert optimization. Because supervised fine-tuning (SFT) supervises single-step denoising at random noise levels whereas during inference the policy iteratively refines self-generated states from uniform noise over a fixed schedule, online RL is critical for calibrating test-time denoising trajectories~\cite{dppo,dvlarl}. Unlike masked diffusion RL~\cite{dvlarl}---which omits model-dependent Gumbel-TopK token-selection probabilities and only updates newly unmasked subsets---or endpoint-reconstruction methods~\cite{udmgrpo}, metric DFM updates the full sequence via a parameter-independent metric transition kernel conditioned on clean proposals. Formulating inference on a \textit{proposal-augmented stochastic path} that records both noisy states and sampled clean proposals allows the transition kernels to cancel out \textit{exactly} in the PPO~\cite{ppo} importance ratio. Finally, to move beyond sparse task rewards, we predict latent visual changes strictly from current context~\cite{miao2026internw0delta} and reward consistency between predicted and observed transitions after each action chunk to jointly optimize both experts.

In summary, our main contributions are threefold:
\begin{itemize}
    \item \textbf{We show that a unified discrete WAM architecture and training paradigm can match or surpass much larger continuous WAMs.} With only 2.3B parameters, DFM-WAM achieves 95.35\% on LIBERO~\cite{libero} without embodied pretraining, and scales with embodied pretraining to 91.47\% (Full) and 54.74\% (Clean-to-Random, C2R) average success across clean and randomized scenes on 50-task RoboTwin 2.0~\cite{robotwin}, matching or surpassing 5B--8B continuous WAMs~\cite{fastwam,lingbotva,motus}.
    \item \textbf{We propose an RL method that derives exact path-level importance sampling ratios for multi-step DFM denoising.} This enables PPO~\cite{ppo} to directly calibrate the policy under the test-time denoising schedule, elevating LIBERO~\cite{libero} average success from 95.35\% to 99.7\% and RoboTwin eight-task success from 83.25\% to 92.97\%.
    \item \textbf{We introduce a mixed-reward RL method combining dense visual-change rewards with sparse task rewards.} Rewarding consistency between predicted and observed visual changes alongside sparse task rewards allows a \textit{single PPO loss} to jointly optimize both Video and Action experts without explicit future-frame generation, pushing RoboTwin eight-task success to 93.75\%.
\end{itemize}

%% file: sec/2_related_work.tex
\section{Related Work}

\begin{figure*}[t]
  \centering
  \includegraphics[width=\linewidth]{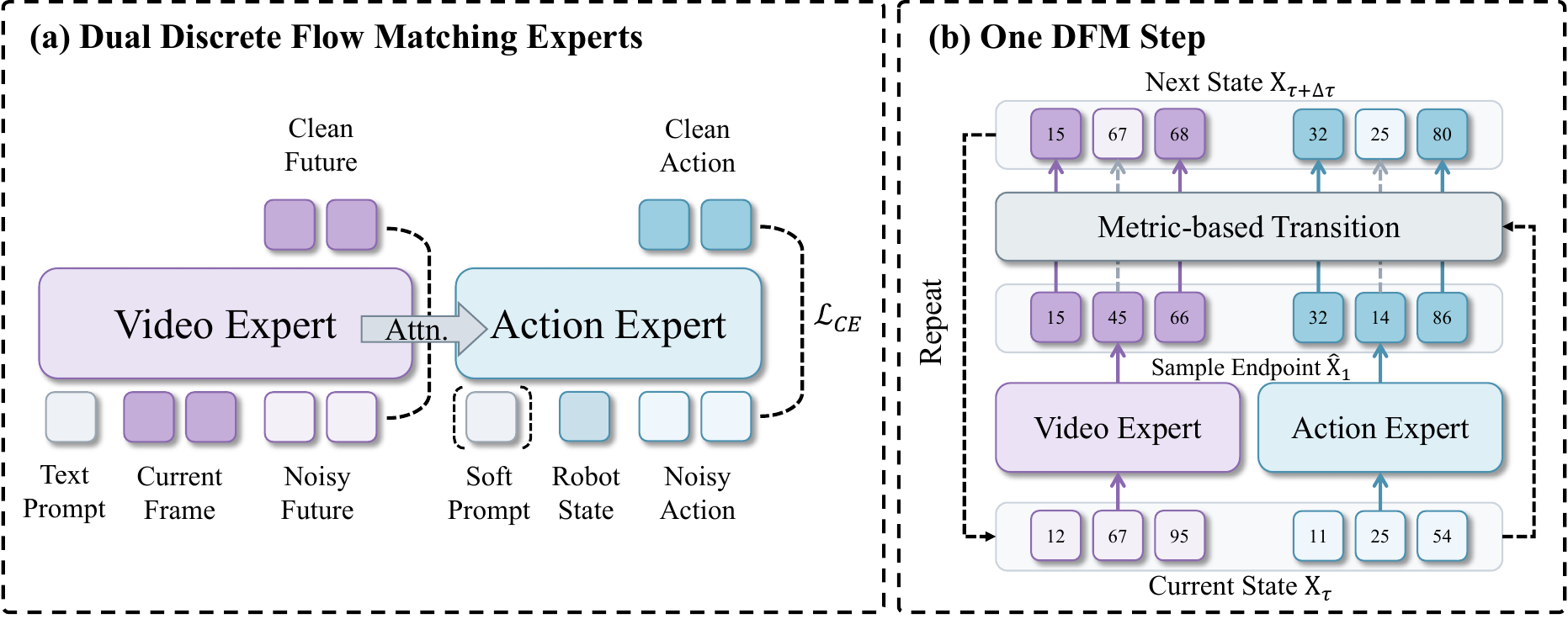}
  \caption{\textbf{Base \method architecture and discrete generation.} (a) The Video and Action Experts predict future head-frame tokens and an action chunk, with actions conditioned on the state and current-context keys and values; future-frame prediction is used only during training, while inference reuses cached context for action-only denoising. (b) One intermediate stochastic DFM step for both modalities. A sampled clean endpoint guides a modality-specific metric transition that either changes a token or retains its current value. The next state need not equal the sampled endpoint.}
  \label{fig:pipeline}
\end{figure*}

\textbf{World-action modeling from generative priors.}
Generative policies effectively capture multimodal action distributions for closed-loop control, as pioneered by Diffusion Policy~\cite{diffusionpolicy} and flow-based VLA models such as $\pi_0$~\cite{pi0}, $\pi_{0.5}$~\cite{pi05}. Beyond encoding current observations, WAMs internalize physical dynamics by predicting future visual observations~\cite{vidar,geact,geact2}. Recent studies further clarify how visual generation benefits control and efficiency~\cite{efficientwam}: Fast-WAM~\cite{fastwam} and Simple-WAM~\cite{simplewam} decouple training-time future supervision from test-time imagination, while ImageWAM~\cite{imagewam} and PatchWAM~\cite{patchwam} demonstrate that supervising single-frame or patch-level future observations provides rich dynamics priors without multi-frame rollout overhead. However, these high-performing WAMs predominantly rely on continuous diffusion or flow backbones~\cite{cosmospolicy,lingbotva,fastwam,motus,imagewam}. DFM-WAM demonstrates that a compact 2.3B discrete WAM built on a pretrained discrete video backbone and trained with parallel future-frame and action DFM rivals 5B--8B continuous WAMs across embodied pretraining and downstream control.

\textbf{Discrete world and action generation.}
Discrete representations offer a unified token space for multimodal modeling~\cite{openvla,pi0fast,univla} and exhibit strong state reactivity without the temporal over-smoothing of continuous regression~\cite{yi2026discrete}. For world modeling, VPDD~\cite{vpdd} transfers actionless discrete video pretraining to guide a separate action denoiser, while WorldVLA~\cite{worldvla} autoregressively decodes visual and action tokens at the cost of high inference latency. To enable parallel decoding, MMaDA-VLA~\cite{mmadavla}, SUAVE~\cite{suave}, MM-ACT~\cite{mmact}, and DD-VLA~\cite{ddvla} adopt masked discrete diffusion (DD). However, masked DD suffers from an \textit{irreversible commitment} problem---under standard monotonic unmasking schedules, generated tokens remain fixed in subsequent steps~\cite{dfmvla}. While recent studies~\cite{dfmvla,yi2026discrete} show that DFM~\cite{dfm} mitigates both irreversible commitment and cross-dimension factorization errors via iterative full-sequence action refinement, they condition solely on static current observations without modeling future visual transitions. Building upon metric-path discrete video generation in URSA~\cite{ursa} over FSQ~\cite{fsq} lattices, DFM-WAM unifies both future-frame prediction and iterative action refinement under metric-path DFM.

\textbf{Reinforcement learning over discrete generation paths.}
Calibrating iterative generative policies under fixed few-step test-time denoising motivates online RL across internal generation trajectories~\cite{ppo,dppo,dvlarl}. In masked DD, dVLA-RL~\cite{dvlarl} optimizes action-denoising trajectories in MM-ACT~\cite{mmact} by factorizing transitions into confidence-driven Gumbel-TopK token selection ($P_{\text{unmask}}$) and token generation over newly unmasked subsets; however, because evaluating $P_{\text{unmask}}$ is intractable, it omits model-dependent token-selection factors in the policy ratio and only updates unmasked subsets at each step. For DFM, DoMinO~\cite{domino} formulates inference as an internal Markov decision process using jump and stay probabilities, whereas UDM-GRPO~\cite{udmgrpo} reconstructs noisy states from generated endpoints to bypass instability with naive intermediate proposals. DFM-WAM highlights a key probabilistic advantage of metric DFM over masked DD: because state transitions are governed by a parameter-independent metric kernel conditioned on full-sequence clean proposals, formulating inference on a \textit{proposal-augmented path} allows transition kernels to cancel out \textit{exactly} in the PPO~\cite{ppo} ratio---enabling exact path-level likelihood ratios across the full sequence without omitting selection factors or relying on endpoint reconstruction approximations.

\textbf{Visual feedback for world and policy improvement.}
Visual rewards provide dense supervision for post-training world models and policies. RLVR-World~\cite{rlvrworld} and Future-Worlds~\cite{futureworlds} refine action-conditioned world predictors using perceptual or motion-aware rewards without updating the control policy, whereas VLA-RFT~\cite{vlarft} freezes an autoregressive world simulator to reward a continuous action head. Closer to joint adaptation, WAM-RL~\cite{wamrl} combines prediction-execution reconstruction rewards with separate online video SFT passes, and GigaBrain-0.5M*~\cite{ramp} couples world-value learning with advantage-conditioned policy fitting. Rather than relying solely on sparse task rewards~\cite{dppo,dvlarl} or requiring separate world-model SFT objectives and explicit future-frame generation during RL~\cite{wamrl}, DFM-WAM adapts the Causal Imprint mechanism from InternW0-$\Delta$~\cite{miao2026internw0delta} into Visual Change Queries (VCQs) to construct a dense latent visual-change consistency reward, enabling a single PPO~\cite{ppo} objective to jointly update both experts via action-proposal gradients.

%% file: sec/3_method.tex
\section{Method}

DFM-WAM unifies future visual modeling and visuomotor control within a dual-expert DFM~\cite{dfm} paradigm. By decoupling dynamics representation from test-time execution via an asymmetric dual-expert architecture (Sec.~\ref{sec:arch}), we supervise parallel future-frame and action generation along ordinal-preserving metric paths during pretraining and SFT without incurring inference-time visual rollout latency (Sec.~\ref{sec:sft}). Crucially, because multi-step DFM transitions are governed by a parameter-independent metric kernel conditioned on clean endpoint proposals, this generative formulation naturally enables policy optimization with exact path-level ratios and dense visual-change credit assignment to jointly optimize both experts during online RL (Sec.~\ref{sec:rl}).

\subsection{Overall Architecture}
\label{sec:arch}

At control step $t$, given a language instruction $l$, current visual observations $o_t$, and proprioceptive robot state $s_t$, DFM-WAM predicts an action chunk $\mathbf{a}_t = (a_t, \dots, a_{t+H-1})$ of horizon $H$ alongside a future head-camera observation $o_{t+H}^{\text{head}}$. As illustrated in Fig.~\ref{fig:pipeline}(a), DFM-WAM pairs a 1.7B Video Expert initialized from URSA-1.7B~\cite{ursa} with a 0.6B Action Expert whose transformer blocks are initialized from URSA-0.6B with newly initialized control interfaces, forming a compact 2.3B model that transfers discrete video-generation priors to visuomotor control.

Visual observations are encoded into discrete tokens via a frozen FSQ image tokenizer~\cite{cosmostokenizer}, which replaces URSA's original video tokenizer to provide a compact spatial representation for sparse current and future frames. Continuous robot states and actions are uniformly quantized into discrete bins. The two experts maintain separate parameters and interact through directed joint self-attention. Following the efficient conditioning design of Fast-WAM~\cite{fastwam}, action tokens attend to the current visual-language context, robot state, and optional task/embodiment soft prompts, while future visual tokens form a parallel prediction branch strictly masked from the causal path of action generation. This directed attention design allows training-time future-frame supervision to shape the visual backbone while enabling fast, action-only test-time inference using cached context keys and values.

Both modalities generate sequences by iteratively refining categorical noise over discrete steps (Fig.~\ref{fig:pipeline}(b)). At each intermediate stochastic step, each expert predicts a categorical distribution over clean target tokens and samples a clean endpoint proposal. A modality-specific metric transition kernel then determines token updates, allowing each token to either jump toward the sampled endpoint or retain its current state~\cite{ursa}. Detailed tokenization, attention masks, and benchmark interfaces are provided in Appendix~\ref{sec:supp-model}.

\subsection{Pretraining and Supervised Fine-Tuning}
\label{sec:sft}
Both experts are trained to denoise corrupted discrete sequences along a metric probability path~\cite{ursa}. For modality $m \in \{v, a\}$, let $x_1^m$ denote a clean target token and $\tau \in [0, 1)$ denote the flow time. The conditional corruption distribution is defined as:
\begin{equation}
    p_\tau^m(x \mid x_1^m) = \frac{\exp[-\beta_m(\tau) d_m(x, x_1^m)]}{\sum_u \exp[-\beta_m(\tau) d_m(u, x_1^m)]},
\label{eq:metric_path}
\end{equation}
where $\beta_m(\tau) = c_m (\tau / (1-\tau))^{\alpha_m}$ monotonically increases from $0$ to $\infty$. Starting from a uniform prior at $\tau=0$, increasing $\tau$ concentrates probability mass on tokens closer to $x_1^m$ under the modality-specific distance $d_m(\cdot, \cdot)$. Crucially, $d_v$ measures the squared Euclidean distance between normalized FSQ lattice coordinates, while $d_a$ measures the squared difference between normalized action bin indices, preserving ordinal structure in both spaces. During training, we independently sample flow times for vision and action, corrupt the clean targets via Eq.~\eqref{eq:metric_path}, and optimize the joint cross-entropy objective over clean token predictions:
\begin{equation}
    \mathcal{L}_{\text{sup}} = \lambda_v \mathcal{L}_{\text{vid}} + \lambda_a \mathcal{L}_{\text{act}},
\label{eq:sup_loss}
\end{equation}
where $\mathcal{L}_{\text{vid}}$ and $\mathcal{L}_{\text{act}}$ are computed over future head-frame tokens and valid action tokens, respectively, with $\lambda_v = \lambda_a = 1$ during joint training. Because action tokens attend to the Video Expert's context representations, $\mathcal{L}_{\text{act}}$ also backpropagates gradients into the Video Expert.

\begin{figure}[t]
  \centering
  \includegraphics[width=0.9\linewidth]{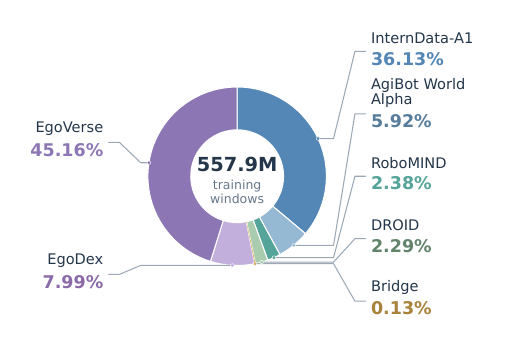}
  \caption{Pretraining corpus composition over 557.9M windows spanning human egocentric videos and simulated/real robot data.}
  \label{fig:pretrain-data}
\end{figure}

\textbf{Large-scale embodied pretraining.}
To investigate how discrete flow world-action modeling scales with heterogeneous data, we curate a pretraining corpus of 557.9M training windows combining action-free human egocentric videos (EgoVerse~\cite{egoverse}, EgoDex~\cite{egodex}) with simulated and real-world robot demonstrations (InternData-A1~\cite{interndataa1}, AgiBot World Alpha~\cite{agibotalpha}, RoboMIND~\cite{robomind}, DROID~\cite{droid}, and Bridge~\cite{bridgedata}; see Fig.~\ref{fig:pretrain-data}). Human videos provide future-visual supervision ($\mathcal{L}_{\text{vid}}$), while robot trajectories supervise both modalities ($\mathcal{L}_{\text{vid}} + \mathcal{L}_{\text{act}}$). To accommodate diverse robots, we equip the Action Expert with embodiment-specific state/action projection tables, output heads, and learnable embodiment soft-prompt~\cite{zheng2026x} tokens (Fig.~\ref{fig:pipeline}(a)). Pretraining proceeds in a three-stage curriculum: (1) warming up the Video Expert for 80k steps on visual dynamics; (2) adapting the Action Expert for 40k steps with the Video Expert frozen; and (3) jointly optimizing both experts for 200k steps. Downstream SFT then adapts either the generic URSA initialization or this embodied-pretrained checkpoint to target benchmarks.

\begin{figure*}[t]
    \centering
    \includegraphics[width=0.98\textwidth]{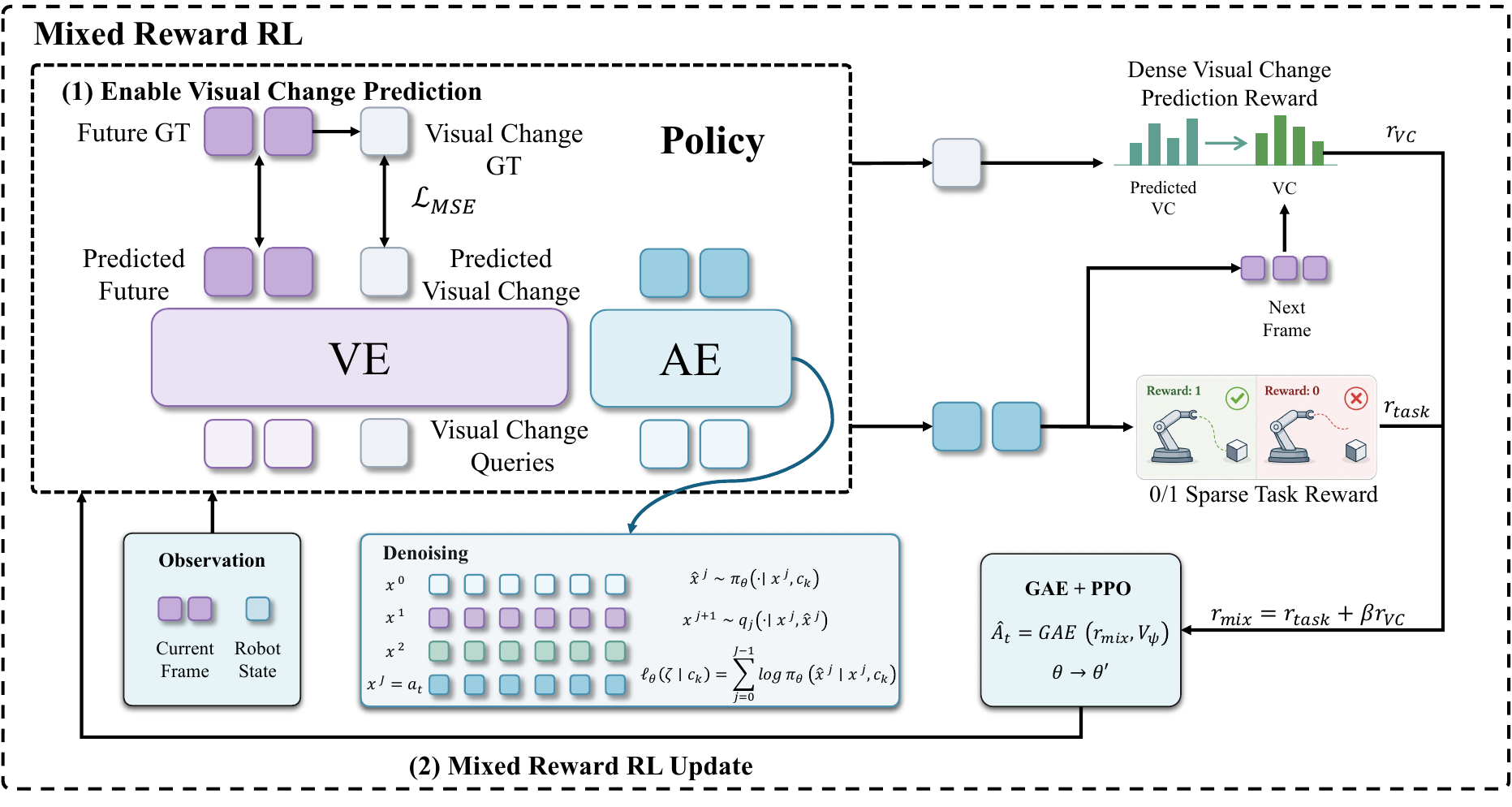}
    \caption{
    \textbf{Overview of reinforcement learning over DFM paths.}
    The Action Expert iteratively generates action proposals through
    multi-step DFM denoising. The parameter-independent metric transition
    kernels cancel in the PPO importance ratio, leaving proposal likelihoods
    for policy optimization. The rollout-derived visual-change consistency
    reward is combined with the sparse task reward and used for the PPO update.
    }
    \label{fig:rl_overview}
\end{figure*}

\subsection{Reinforcement Learning over DFM Paths}
\label{sec:rl}
While SFT supervises single-step denoising across randomly sampled flow times $\tau \in [0, 1)$ corrupted from ground-truth actions, test-time inference omits the future-frame branch and iteratively refines self-generated action states from uniform noise along a fixed $J$-step schedule. Calibrating the policy under this test-time denoising regime via online RL poses a fundamental credit-assignment challenge: an executed action chunk $x^J$ is the marginal result of $J$ sequential stochastic transitions, making exact marginal likelihood evaluation intractable. We resolve this by formulating policy optimization over a \textit{proposal-augmented stochastic path} (Fig.~\ref{fig:rl_overview}).

\textbf{Proposal-augmented trace and exact kernel cancellation.}
At chunk decision step $k$ with context $c_k = (l, o_{t_k}, s_{t_k})$, the Action Expert generates an action chunk of horizon $H$ over $J$ denoising steps without generating future-frame tokens. At denoising step $j \in \{0, \dots, J-1\}$ with current discrete action state $x^j$, the policy $\pi_\theta(\cdot \mid x^j, c_k)$ predicts factorized categorical distributions $\pi_{\theta, hd}$ across horizon positions $h$ and valid action dimensions $d$, from which a clean endpoint proposal $\hat{x}^j$ is sampled; the metric scheduler then draws the next state $x^{j+1} \sim q_j(\cdot \mid x^j, \hat{x}^j)$. Recording both intermediate states and sampled proposals yields the augmented trace $\zeta = (x^{0:J}, \hat{x}^{0:J-1})$ with joint probability:
\begin{equation}
    P_\theta(\zeta \mid c_k) = p(x^0) \prod_{j=0}^{J-1} \left[ \pi_\theta(\hat{x}^j \mid x^j, c_k) \cdot q_j(x^{j+1} \mid x^j, \hat{x}^j) \right].
\label{eq:aug_trace}
\end{equation}
Under a fixed metric scheduler, both the initial noise prior $p(x^0)$ and the transition kernels $q_j(x^{j+1} \mid x^j, \hat{x}^j)$ are independent of $\theta$ and cancel \textit{exactly} in the augmented-path importance ratio:
\begin{equation}
    \frac{P_\theta(\zeta \mid c_k)}{P_{\theta_{\text{old}}}(\zeta \mid c_k)}
    = \prod_{j=0}^{J-1} \frac{\pi_\theta(\hat{x}^j \mid x^j, c_k)}{\pi_{\theta_{\text{old}}}(\hat{x}^j \mid x^j, c_k)}.
\label{eq:exact_ratio}
\end{equation}

\textbf{Locally clipped policy optimization.}
Rather than directly clipping the full-path product in Eq.~\eqref{eq:exact_ratio} across all $J$ denoising steps and $H$ horizon steps---which can suffer from high variance and ratio explosion---we decompose the update into locally clipped ratios at each denoising step $j$ and chunk position $h$ by summing log-probability differences across valid dimensions $\mathcal{D}$:
\begin{equation}
    \begin{aligned}
    \delta_{kjh} &= \sum_{d \in \mathcal{D}} \log \frac{\pi_{\theta, hd}(\hat{x}_{hd}^j \mid x^j, c_k)}{\pi_{\text{old}, hd}(\hat{x}_{hd}^j \mid x^j, c_k)}, \\
    \rho_{kjh} &= \exp(\text{clip}(\delta_{kjh}, -\delta_{\max}, \delta_{\max})).
    \end{aligned}
\label{eq:local_ratio}
\end{equation}
Broadcasting the chunk-level Generalized Advantage Estimate (GAE) $\tilde{A}_k$ across all $(j, h)$ pairs yields our locally clipped PPO-style~\cite{ppo} objective:
\begin{equation}
    \mathcal{L}_{\text{actor}} = -\left\langle \min\left(\rho_{kjh}\tilde{A}_k, \text{clip}(\rho_{kjh}, 1-\epsilon, 1+\epsilon)\tilde{A}_k\right) \right\rangle_w,
\label{eq:ppo_loss}
\end{equation}
where $\langle \cdot \rangle_w$ denotes averaging over valid chunks with inverse trajectory-length weighting, and $\delta_{\max}$ and $\epsilon$ are the log-ratio and PPO clipping thresholds (Appendix~\ref{sec:supp-rl}).

\textbf{Dense visual-change consistency reward via VCQ.}
To move beyond sparse task rewards, we seek dense predictive feedback to guide joint expert optimization. However, explicitly generating future frames during RL rollouts requires costly multi-step Video Expert denoising. Instead, we incorporate learnable VCQs (adapted from \textit{Causal Imprint} in InternW0-$\Delta$~\cite{miao2026internw0delta}) to predict compact latent visual changes in a single pass. Under our directed attention mask, neither VCQs nor the Action Expert access future-frame tokens: VCQs attend only to the current visual-language context, while the Action Expert attends to the current context, robot state, and VCQs. During SFT, VCQs are trained via an auxiliary MSE loss to predict the pre-quantization head-camera latent displacement $\hat{\Delta z}_k$, where $\Delta z_k = z_{k+1} - z_k$ is the observed latent change after executing chunk $k$. During Mixed reward RL, we remove the auxiliary MSE loss and formulate a dense visual-change consistency reward $g_k$ measuring the relative predictive gain over a zero-change baseline:
\begin{equation}
    \begin{aligned}
    g_k &= \frac{\text{MSE}(0, \Delta z_k) - \text{MSE}(\hat{\Delta z}_k, \Delta z_k)}{\max(\text{MSE}(0, \Delta z_k), \epsilon_0)}, \\
    r_k &= r_k^{\text{task}} + \lambda_{\text{VCQ}} \, m_k^{\text{VCQ}} \text{clip}(g_k, -g_{\max}, g_{\max}),
    \end{aligned}
\label{eq:vcq_reward}
\end{equation}
where $\epsilon_0 > 0$ is a small numerical stability constant, $r_k^{\text{task}} \in \{0, 1\}$ is the sparse task-completion reward, $\lambda_{\text{VCQ}}$ weights the dense visual reward clipped to $[-g_{\max}, g_{\max}]$, and $m_k^{\text{VCQ}}$ masks out terminal or truncated chunks (hyperparameter values are detailed in Appendix~\ref{sec:supp-rl}). This combined return drives joint expert optimization solely through the single PPO objective in Eq.~\eqref{eq:ppo_loss}: the Action Expert is rewarded for executing actions that realize the anticipated visual transition, while the Video Expert and VCQs receive direct policy gradients through the conditioning keys and values that parameterize the action proposals. In our Sparse reward RL variant, we optimize the policy using only sparse task rewards (setting $\lambda_{\text{VCQ}} = 0$), optionally freezing the Video Expert for rapid action-only adaptation. We apply this VCQ formulation on RoboTwin eight-task RL, whereas pretraining, LIBERO, and 50-task RoboTwin SFT omit VCQ tokens.

%% file: sec/4_experiments.tex
\section{Experiments}

\subsection{Experimental Setup}
\label{sec:exp_setup}
\textbf{Benchmarks and training protocols.}
We evaluate DFM-WAM on two comprehensive manipulation benchmarks: LIBERO~\cite{libero} and RoboTwin 2.0~\cite{robotwin}. LIBERO comprises four task suites (\textit{Spatial}, \textit{Object}, \textit{Goal}, and \textit{Long}) with 10 tasks each; we train a single shared SFT policy directly from generic URSA initialization across all 40 tasks, followed by suite-specific Sparse reward RL. RoboTwin 2.0 evaluates bimanual manipulation across both a 50-task benchmark and an eight-task RL benchmark following the task split of dVLA-RL~\cite{dvlarl}. The 50-task \textit{Full} setting provides 50 clean and 500 domain-randomized demonstrations per task, whereas the \textit{C2R} setting fine-tunes solely on the 50 clean demonstrations to test zero-shot robustness to randomized scenes; we evaluate both generic URSA initialization and our embodied-pretrained checkpoint on the 50-task benchmark (Sec.~\ref{sec:exp_pretrain}). The eight-task RL setting incorporates VCQs and initializes from a shared SFT checkpoint trained from generic URSA weights on 1,000 randomized demonstrations per task.

\textbf{Evaluation metrics.}
We report mean task success rates (SR, \%) with uniform task weighting. LIBERO SFT policies are evaluated over 50 episodes per task (2,000 total), and 50-task RoboTwin SFT policies use 100 evaluation episodes per task. For online RL, we follow the online evaluation protocol of dVLA-RL~\cite{dvlarl}: on LIBERO, each suite collects 512 rollout trajectories per update across its 10 tasks, and we report the peak suite-level online rollout success rate; on RoboTwin eight-task RL, we record online rollout success across 64 episodes per task (512 total per update) over 4-step stochastic DFM generation and report the peak eight-task mean performance. Following prior works~\cite{fastwam,dvlarl}, external baselines are reported under their official published protocols, while internal ablations are conducted under strictly matched settings. Complete hyperparameter, protocol, ablation, and baseline-source details are provided in Appendices~\ref{sec:supp-model}--\ref{sec:supp-baselines}.

\subsection{Supervised Control from Discrete Video Priors}
\label{sec:exp_sft}
Table~\ref{tab:exp-sft} benchmarks DFM-WAM against leading continuous and discrete VLAs and WAMs prior to RL post-training. Across both benchmarks, DFM-WAM demonstrates strong parameter efficiency and control accuracy:

\begin{table*}[t]
  \centering
  \small
  \setlength{\tabcolsep}{2.5pt}
  \renewcommand{\arraystretch}{1.08}
  \caption{SFT SR (\%). LIBERO Avg. is the mean of four suites. C2R gives clean- and randomized-scene SR after clean-only SFT. Baseline variants and source tables are detailed in Appendix~\ref{sec:supp-baselines}.}
  \label{tab:exp-sft}
  \begin{tabular}{@{}lcc*{9}{c}@{}}
    \toprule
    & & & \multicolumn{5}{c}{LIBERO} & \multicolumn{2}{c}{RoboTwin Full} & \multicolumn{2}{c}{RoboTwin C2R} \\
    \cmidrule(lr){4-8}\cmidrule(lr){9-10}\cmidrule(l){11-12}
    Method & \shortstack{Continuous/\\Discrete} & \shortstack{Params.\\(B)} & Spat. & Obj. & Goal & Long & Avg. & Clean & Random & Clean & Random \\
    \midrule
    \multicolumn{12}{@{}l}{\textbf{VLA}} \\
    $\pi_0$~\cite{pi0} & Continuous & 3.3 & 96.8 & 98.8 & 95.8 & 85.2 & 94.15 & 65.92 & 58.40 & 46.42 & 16.34 \\
    $\pi_{0.5}$~\cite{pi05} & Continuous & 3.3 & 98.8 & 98.2 & 98.0 & 92.4 & 96.85 & 82.74 & 76.76 & 70.66 & 46.00 \\
    OpenVLA-OFT~\cite{openvlaoft} & Continuous & 7 & 97.6 & 98.4 & 97.9 & 94.5 & 97.10 & --- & --- & --- & --- \\
    $\pi_0$-FAST~\cite{pi0fast} & Discrete & 3 & 96.4 & 96.8 & 88.6 & 60.2 & 85.50 & --- & --- & --- & --- \\
    OpenVLA~\cite{openvla} & Discrete & 7 & 84.7 & 88.4 & 79.2 & 53.7 & 76.50 & --- & --- & --- & --- \\
    DFM-VLA~\cite{dfmvla} & Discrete & 8.5 & 96.8 & 98.8 & 94.4 & 92.6 & 95.65 & --- & --- & --- & --- \\
    DD-VLA~\cite{ddvla} & Discrete & 7 & 97.2 & 99.4 & 96.8 & 92.2 & 96.40 & --- & --- & --- & --- \\
    MM-ACT~\cite{mmact} & Discrete & 8 & 97.8 & 99.4 & 94.8 & 93.0 & 96.25 & --- & --- & --- & --- \\
    \midrule
    \multicolumn{12}{@{}l}{\textbf{WAM}} \\
    Fast-WAM~\cite{fastwam} & Continuous & 6 & 98.2 & 100.0 & 97.0 & 95.2 & 97.60 & 91.88 & 91.78 & 77.76 & 1.92 \\
    Cosmos Policy~\cite{cosmospolicy} & Continuous & 2 & 98.1 & 100.0 & 98.2 & 97.6 & 98.48 & --- & --- & --- & --- \\
    LingBot-VA~\cite{lingbotva} & Continuous & 5.3 & 98.5 & 99.6 & 97.2 & 98.5 & 98.45 & 92.93 & 91.55 & --- & --- \\
    Motus~\cite{motus} & Continuous & 8 & 96.8 & 99.8 & 96.6 & 97.6 & 97.70 & 88.66 & 87.02 & --- & --- \\
    ImageWAM~\cite{imagewam} & Continuous & 4.6 & 97.2 & 99.2 & 98.8 & 98.4 & 98.40 & 93.20 & 93.56 & --- & --- \\
    WorldVLA~\cite{worldvla} & Discrete & 7 & 87.6 & 96.2 & 83.4 & 60.0 & 81.80 & --- & --- & --- & --- \\
    MMaDA-VLA~\cite{mmadavla} & Discrete & 8 & 98.8 & 99.8 & 98.0 & 95.2 & 97.95 & --- & --- & --- & --- \\
    \midrule
    \textbf{DFM-WAM} & Discrete & 2.3 & 96.8 & 98.2 & 95.8 & 90.6 & 95.35 & 91.92 & 91.02 & 77.84 & 31.64 \\
    \bottomrule
  \end{tabular}
\end{table*}

\begin{itemize}
    \item \textbf{Parameter efficiency among discrete baselines:} With only 2.3B total parameters, DFM-WAM achieves 95.35\% average success on LIBERO, outperforming the 7B autoregressive discrete world model WorldVLA~\cite{worldvla} by \textbf{+13.55\%} (and by \textbf{+30.6\%} on long-horizon tasks) as well as the 3B discrete policy $\pi_0$-FAST~\cite{pi0fast} by \textbf{+9.85\%}. It is also comparable to much larger 7B--8.5B discrete diffusion VLAs such as DFM-VLA (95.65\%)~\cite{dfmvla}, DD-VLA (96.40\%)~\cite{ddvla}, and MM-ACT (96.25\%)~\cite{mmact} at less than one-third of their parameter count.
    \item \textbf{Competitiveness against larger continuous WAMs:} On the challenging 50-task bimanual RoboTwin Full benchmark, DFM-WAM achieves \textbf{91.92\%} in clean scenes and \textbf{91.02\%} in highly randomized scenes. This substantially surpasses the 3.3B continuous flow models $\pi_0$ (65.92\%/58.40\%) and $\pi_{0.5}$ (82.74\%/76.76\%), outperforms the 8B continuous world model Motus (88.66\%/87.02\%)~\cite{motus}, and matches the 6B Fast-WAM (91.88\%/91.78\%)~\cite{fastwam} and 5.3B LingBot-VA (92.93\%/91.55\%)~\cite{lingbotva} while using under half the parameters.
\end{itemize}
Controlled ablations in Appendix~\ref{sec:supp-sft-ablations} further confirm the roles of discrete video priors and future-frame supervision: under identical URSA initialization of both experts, joint video--action supervision improves RoboTwin eight-task SFT success from 53.3\% to 86.4\% at 15k steps (Fig.~\ref{fig:supp-rt8-sft-cotrain}).

\subsection{Reinforcement Learning over DFM Paths}
\label{sec:exp_rl}
We next evaluate how our proposal-augmented RL framework calibrates DFM-WAM under the test-time denoising schedule across two paradigms: \textbf{Sparse reward RL ($R_1$)}, which optimizes the policy using sparse task rewards alone, and \textbf{Mixed reward RL ($R_2$)}, which jointly updates both Video and Action experts (along with VCQs) under a \textit{single PPO loss} using sparse task rewards augmented with our dense VCQ visual-change consistency reward (Eq.~\eqref{eq:vcq_reward}).

\textbf{Near-saturated performance on LIBERO.}
As shown in Table~\ref{tab:exp-rl-libero}, starting from our 95.35\% SFT checkpoint, Sparse reward RL consistently improves all four LIBERO suites, elevating average success to \textbf{99.7\%}. Notably, DFM-WAM boosts the challenging \textit{LIBERO-Long} suite by \textbf{+8.8\%} (from 90.6\% to 99.4\%). Overall, DFM-WAM surpasses SimpleVLA-RL (99.1\%) and matches the state-of-the-art 8B dVLA-RL (99.7\%)~\cite{dvlarl} with a significantly more compact 2.3B model.

\begin{table}[t]
  \centering
  \small
  \setlength{\tabcolsep}{2.5pt}
  \caption{LIBERO SFT and RL SR (\%). Baseline results are from dVLA-RL Table 1~\cite{dvlarl}.}
  \label{tab:exp-rl-libero}
  \begin{tabular}{@{}lccccc@{}}
    \toprule
    Policy & Spat. & Obj. & Goal & Long & Avg. \\
    \midrule
    SimpleVLA-RL & 99.4 & 99.1 & 99.2 & 98.5 & 99.1 \\
    dVLA-RL & 99.8 & \textbf{100.0} & \textbf{99.6} & 99.2 & \textbf{99.7} \\
    \midrule
    \multicolumn{6}{@{}l}{DFM-WAM} \\
    SFT & 96.8 & 98.2 & 95.8 & 90.6 & 95.35 \\
    Sparse reward RL & \textbf{100.0} & \textbf{100.0} & 99.2 & \textbf{99.4} & \textbf{99.7} \\
    \bottomrule
  \end{tabular}
\end{table}

\begin{figure}[t]
  \centering
  \includegraphics[width=\linewidth]{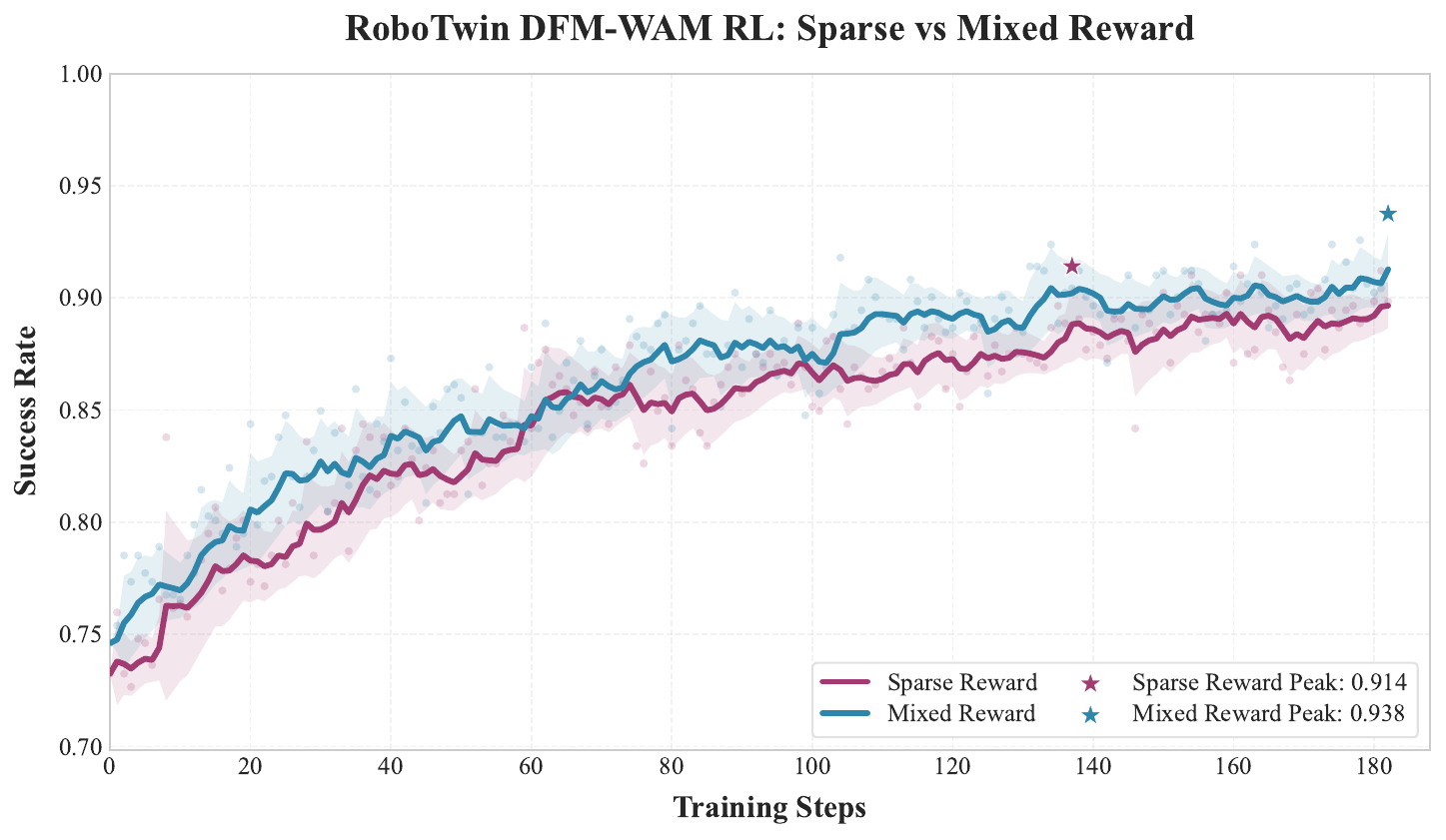}
  \caption{\textbf{RoboTwin DFM-WAM RL success rate comparison under sparse and mixed rewards.} The purple curve represents action-only training with the sparse task reward, whereas the blue curve represents jointly training both Video and Action experts with a mixed reward that combines the sparse task reward with the dense visual-change consistency reward via VCQs. Shaded regions indicate exponential moving average (EMA) $\pm$ exponentially weighted temporal dispersion computed from historical training steps. The curves are plotted up to step 183, where the mixed-reward run reaches its peak of 93.75\% (the sparse-reward run reaches its full-run peak of 92.97\% at step 215).}
  \label{fig:exp-rl-curves}
\end{figure}

\textbf{Joint RL with dense visual feedback on RoboTwin.}
Table~\ref{tab:exp-rl-robotwin} and Fig.~\ref{fig:exp-rl-curves} present results on the eight-task randomized RoboTwin benchmark. Even before RL, our SFT baseline ($R_0$: 83.25\%) substantially outperforms continuous SFT references $\pi_{0.5}$ (74.9\%) and OpenVLA-OFT (34.2\%), as well as SimpleVLA-RL (68.7\%). Applying action-only Sparse reward RL ($R_1$, with the Video Expert frozen) over our proposal-augmented path calibrates the policy under the test-time denoising schedule and lifts online rollout success from an initial stochastic baseline of 73.24\% to \textbf{92.97\%}, outperforming the 8B dVLA-RL baseline (92.0\%). Furthermore, unfreezing the Video Expert and incorporating the dense latent visual-change reward in Mixed reward RL ($R_2$) pushes peak performance further to \textbf{93.75\%} without requiring explicit future-frame generation or separate video SFT passes. As shown in Table~\ref{tab:exp-rl-robotwin} and Fig.~\ref{fig:exp-rl-curves}, $R_2$ reaches a higher observed peak success rate than $R_1$ earlier in training (93.75\% at update 183 versus 92.97\% at update 215 for $R_1$). Within the common 183-update window, adding the dense visual-change reward improves the observed peak under both action-only and joint expert updates (Fig.~\ref{fig:supp-reward-scope}).

\begin{table}[t]
  \centering
  \small
  \setlength{\tabcolsep}{2.5pt}
  \caption{RoboTwin eight-task SFT and RL SR (\%). Baselines are from dVLA-RL Table 2~\cite{dvlarl}; $\pi_{0.5}$ and OpenVLA-OFT are SFT references. Task-level results are in Table~\ref{tab:supp-rt8-tasks}.}
  \label{tab:exp-rl-robotwin}
  \begin{tabular}{@{}lcc@{}}
    \toprule
    Policy & Task setting & Avg. \\
    \midrule
    $\pi_{0.5}$ & Multitask & 74.9 \\
    OpenVLA-OFT & Single-task & 34.2 \\
    SimpleVLA-RL & Single-task & 68.7 \\
    dVLA-RL & Multitask & 92.0 \\
    \midrule
    \multicolumn{3}{@{}l}{DFM-WAM} \\
    $R_0$: SFT & Multitask & 83.25 \\
    $R_1$: Sparse reward RL & Multitask & 92.97 \\
    $R_2$: Mixed reward RL & Multitask & \textbf{93.75} \\
    \bottomrule
  \end{tabular}
\end{table}

\subsection{Scaling with Heterogeneous Embodied Pretraining}
\label{sec:exp_pretrain}
To assess how our unified discrete WAM benefits from large-scale cross-embodiment data, Table~\ref{tab:exp-transfer} compares 50-task RoboTwin SFT models initialized directly from generic URSA video weights against those initialized from our checkpoint pretrained on the 557.9M-window corpus. Across all 50 bimanual tasks in the \textit{Full} setting, large-scale embodied pretraining improves average evaluation success from 88.46\% to \textbf{91.92\%} in clean environments and from 88.26\% to \textbf{91.02\%} under domain randomization. Moreover, in the \textit{C2R} setting, embodied pretraining boosts performance from 57.68\% to \textbf{77.84\%} (clean) and from 2.52\% to \textbf{31.64\%} (randomized), demonstrating that heterogeneous embodied pretraining improves both downstream control and robustness to scene randomization (see Appendix~\ref{sec:supp-results} for per-task breakdowns, additional ablations, and qualitative future-frame generations).

\begin{table}[t]
  \centering
  \small
  \caption{Pretraining transfer on 50 RoboTwin tasks (SR, \%). Embodied PT marks initialization from embodied pretraining on our 557.9M-window corpus (versus direct URSA initialization).}
  \label{tab:exp-transfer}
  \begin{tabular}{@{}lccc@{}}
    \toprule
    SFT data & Embodied PT & Clean & Random \\
    \midrule
    Full & $\times$ & 88.46 & 88.26 \\
    Full & $\checkmark$ & \textbf{91.92} & \textbf{91.02} \\
    C2R & $\times$ & 57.68 & 2.52 \\
    C2R & $\checkmark$ & \textbf{77.84} & \textbf{31.64} \\
    \bottomrule
  \end{tabular}
\end{table}

\subsection{Real-World Experiments}
\label{subsec:real_world}

We evaluate \method on the dual-arm ALOHA platform~\cite{zhao2023learningfinegrainedbimanualmanipulation} using a long-horizon task, \textit{Organize Table}. The task comprises three sequential stages: transferring a carrot from a plate into a pot (PnP Carrot), lifting and moving the pot with both arms to reveal a coaster underneath (Lift Pot), and placing a cup onto the coaster (Place Empty Cup). We collect 150 demonstrations and fine-tune DFM-WAM and the $\pi_{0.5}$ baseline on them. We evaluate each policy over 50 trials. Further details are provided in Appendix~\ref{sec:supp-real-world}.

\begin{table}[t]
\centering
\caption{Real-world performance (\%) on ALOHA. Avg. is the average cumulative stage success rate across the three \textit{Organize Table} stages. Column-best results are bold.}
\label{tab:real_world_main}
\small
\setlength{\tabcolsep}{4pt}
\renewcommand{\arraystretch}{1.08}
\begin{tabular}{@{}lcccc@{}}
\toprule
Method & \multicolumn{3}{c}{Organize Table} & Avg. \\
\cmidrule(lr){2-4}
 & PnP Carrot & Lift Pot & Place Empty Cup & \\
\midrule
$\pi_{0.5}$ & 88.0 & 70.0 & 68.0 & 75.3 \\
\textbf{DFM-WAM} & \textbf{94.0} & \textbf{80.0} & \textbf{74.0} & \textbf{82.7} \\
\bottomrule
\end{tabular}
\end{table}

As shown in Table~\ref{tab:real_world_main}, DFM-WAM consistently outperforms $\pi_{0.5}$ across all three stages of the real-world Organize Table task, increasing the average stage success rate from 75.3\% to 82.7\%. The largest improvement is observed on Lift Pot, which requires coordinated dual-arm manipulation. These results demonstrate the effectiveness of DFM-WAM in the real world across multiple stages of a long-horizon manipulation task, including object transfer, bimanual manipulation, and precise placement.

%% file: sec/5_conclusion.tex
\section{Conclusion}

We introduced \textbf{DFM-WAM}, a compact 2.3B discrete world-action model unifying future-frame modeling and visuomotor control via metric-path DFM. Across pretraining, SFT, and online RL, our study establishes three findings:
(1) generic discrete video priors transfer directly to control (95.35\% on LIBERO without embodied pretraining), while scaling parallel video-action DFM on a corpus of 557.9M heterogeneous windows boosts 50-task RoboTwin success to 91.92\%/91.02\% (Full) and 77.84\%/31.64\% (C2R)---matching 5B--8B continuous WAMs and outperforming $\pi_{0.5}$ on real-world ALOHA (82.7\% vs.\ 75.3\% average cumulative stage SR);
(2) formulating multi-step DFM inference on a proposal-augmented path yields exact augmented-path importance ratios via metric-kernel cancellation, elevating LIBERO success to 99.7\% and RoboTwin eight-task success from 83.25\% to 92.97\%; and
(3) combining dense visual-change consistency rewards with sparse task rewards allows a single PPO objective to jointly optimize both Video and Action experts without explicit future-frame generation, reaching 93.75\% on RoboTwin.

%% file: sec/supp_model.tex
\section{Model Architecture and Discrete Generation}
\label{sec:supp-model}

\subsection{Expert initialization and control interface}
The Video and Action Experts have separate weights and share an attention-head layout (Table~\ref{tab:supp-experts}). At each layer, queries from each expert attend over the permitted keys and values from both streams. The Video Expert inherits URSA's visual embedding, output head, and token metric~\cite{ursa}. The Action Expert inherits the transformer blocks and final normalization of the smaller URSA model. Its state and action embedding tables and action output head are newly initialized for robot control.

\begin{table}[htb]
\centering\small
\caption{Expert dimensions and initialization. The initialization row names the source URSA checkpoints.}
\label{tab:supp-experts}
\begin{tabular}{@{}ccc@{}}
\toprule
 & Video Expert & Action Expert\\
\midrule
Initialization & URSA-1.7B & URSA-0.6B\\
Layers & 28 & 28\\
Hidden / MLP width & 2,048 / 6,144 & 1,024 / 3,072\\
Query / key-value heads & 16 / 8 & 16 / 8\\
Dimension per head & 128 & 128\\
\bottomrule
\end{tabular}
\end{table}

\paragraph{Visual representations.}
The frozen Cosmos DI16$\times$16 image tokenizer encodes frames independently in FP32~\cite{cosmostokenizer}. Its six FSQ coordinates have $(8,8,8,5,5,5)$ levels, yielding 64,000 possible codes~\cite{fsq}. URSA's original video tokenizer compresses four frames in time and eight pixels along each spatial axis. The image tokenizer shares its code-index ordering and supports direct weight transfer by index. Its larger spatial stride reduces the number of tokens for a $256\times320$ head image from 1,280 to 320.

\paragraph{RoboTwin control.}
The policy observes a head image ($256\times320$), two wrist images ($128\times160$ each), the current joint state, and a task instruction. Together the current images produce 480 tokens. The eight-task instruction is padded to 128 tokens and specifies the embodiment, control representation, data rate, and task. The supervised targets are 16 actions and the head image 16 steps ahead. For each arm joint, the target at offset $h$ is $q^\star_{t+h}-q_t$, so all 16 offsets share one anchor. Gripper targets remain absolute. We normalize the state and action using separate per-dimension min--max statistics, clip to $[-1,1]$, and quantize to 2,048 bins. The eight-task interface places 14 valid dimensions in 16 slots per step and fixes the two padding slots to bin 1,024. Decoding inverts the normalization at the selected bin centers and restores the anchor for arm joints. The LIBERO policy uses seven action dimensions and receives no robot-state input.

\subsection{Directed attention and cached inference}
\begin{figure*}[t]
\centering
\includegraphics[width=\textwidth]{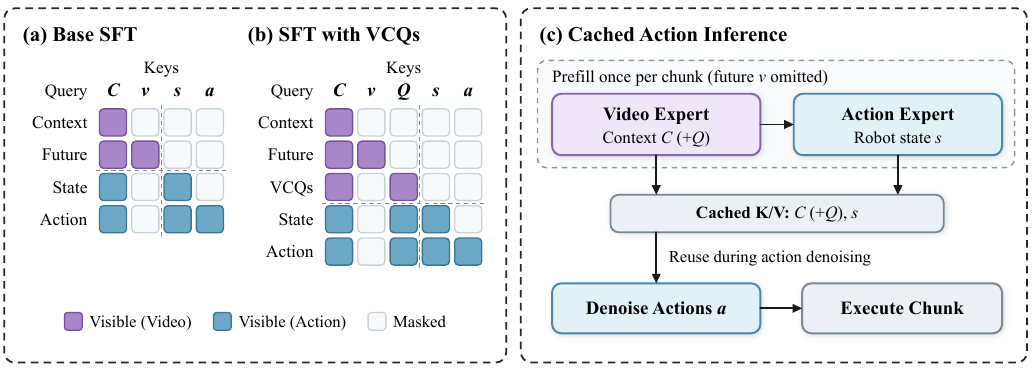}
\caption{\textbf{Directed attention and cached inference.} (a,b) A filled cell lets the row's query read the column's keys, shaded by the querying expert (Video or Action). Only future-frame queries read $v$; with VCQs, state and action also read $Q$. (c) Action inference omits $v$, prefills context and state keys and values (K/V) once per chunk, and reuses the cached K/V during action denoising.}
\label{fig:supp-attention}
\end{figure*}

Figure~\ref{fig:supp-attention} shows the directed attention used in RoboTwin 8 tasks, following Fast-WAM~\cite{fastwam}. $C$ groups the instruction, visual delimiter, and current views. The other groups contain the noisy future head frame $v$, optional VCQs $Q$, robot state $s$, and noisy action chunk $a$. Within each group, attention is bidirectional.

At each new observation, the Video Expert caches $C$ and, when present, $Q$. The Action Expert reads these features to compute the state cache. Denoising then updates only the action tokens. Both experts use shared 3D rotary coordinates, with wrist views below the head view. State and action tokens keep their training positions when $v$ is omitted. After executing up to 16 actions, the policy observes again and rebuilds both caches.

\subsection{Metric path and finite-step sampler}
Following Eq.~\eqref{eq:metric_path}, we adopt URSA's metric-path corruption schedule and finite-step discrete flow sampler~\cite{ursa} (shift~4, $\alpha_v=\alpha_a=1$, $c_v=5$, and $c_a=3$), including its numerical boundary conventions. The video distance $d_v$ is the squared Euclidean distance on L2-normalized FSQ lattice coordinates, and the action distance over 2,048 bins is $d_a(i,l)=((i-l)/2047)^2$. During supervised training, each modality independently draws $u\sim\mathcal U(0,1)$ with $\sigma=1-0.999u$ and $\tau=1-4\sigma/(1+3\sigma)$. During $J$-step action inference, we use the shift-4 schedule $\sigma_j=1-j/J$ and $\tau_j=1-4\sigma_j/(1+3\sigma_j)$ ($j\in\{0,\dots,J-1\}$). Each valid action token is initialized uniformly over the 2,048 bins; at each step $j$, a cached Action Expert forward pass predicts clean-endpoint logits to obtain a proposal $\hat x^j$, followed by URSA's parameter-independent metric transition kernel $q_j(x^{j+1}\mid x^j,\hat x^j)$ for $j<J-1$ and direct clean-endpoint readout $x^J=\hat x^{J-1}$ at the final step. SFT evaluation selects the final proposal greedily by argmax, whereas online RL rollouts sample proposals stochastically at all $J$ steps to maintain full categorical support over the proposal-augmented path (Eq.~\eqref{eq:aug_trace}).

%% file: sec/supp_pretraining.tex
\section{Pretraining Data and Supervised Training}
\label{sec:supp-pretrain}

\subsection{Corpus construction}
A training window contains a current observation and a head-camera target 16 steps later. Robot windows also provide the current state and the intervening 16 actions. Overlapping windows can share frames. The corpus contains 261,397,585 robot training windows and 296,530,527 video-only training windows. Episodes contributing windows span approximately 8,800 hours across the training and validation splits, counting each source recording once across temporal strides.

\begin{table*}[t]
  \centering
  \caption{Pretraining inventory and expected sampling probabilities. Each robot row specifies one mapping from a source robot's state, actions, and cameras to model inputs and targets. Action space gives the arm coordinate system. Action dimension counts valid action coordinates, including grippers. Window counts are in millions. The three rightmost columns give the expected percentage of all samples drawn during each stage.}
  \label{tab:supp-pt-inventory}
  \small
  \setlength{\tabcolsep}{2.5pt}
  \begin{tabular*}{\textwidth}{@{\extracolsep{\fill}}*{9}{c}@{}}
    \toprule
    & & & & \multicolumn{2}{c}{Windows (M)} & \multicolumn{3}{c}{Sampling probability (\%)} \\
    \cmidrule(lr){5-6}\cmidrule(l){7-9}
    Source & Mapping & Action space & Action dim. & Train & Val. & Video only & Action only & Joint \\
    \midrule
    InternData-A1~\cite{interndataa1} & Franka & Joint & 8 & 40.648 & 0.129 & 7.70 & 15.40 & 12.32 \\
    & Genie-1 & Joint & 16 & 10.220 & 0.121 & 2.73 & 5.47 & 4.38 \\
    & Lift-2 & Joint & 14 & 76.724 & 0.390 & 12.40 & 24.81 & 19.85 \\
    & Split Aloha & Joint & 14 & 73.969 & 0.353 & 12.07 & 24.14 & 19.31 \\
    AgiBot World Alpha~\cite{agibotalpha} & A2D & Joint & 16 & 33.026 & 2.958 & 6.59 & 13.18 & 10.55 \\
    RoboMIND~\cite{robomind} & AgileX & End effector & 14 & 5.633 & 0.296 & 1.75 & 3.50 & 2.80 \\
    & Dual Franka FR3 & Joint & 16 & 0.322 & 0.015 & 0.20 & 0.41 & 0.33 \\
    & Franka Panda & Joint & 8 & 1.651 & 0.090 & 0.70 & 1.39 & 1.12 \\
    & Tien Kung Gello & Joint & 16 & 3.542 & 0.181 & 1.24 & 2.47 & 1.98 \\
    & Tien Kung Prod1 & Joint & 16 & 1.407 & 0.069 & 0.62 & 1.24 & 0.99 \\
    & UR5e & Joint & 7 & 0.747 & 0.040 & 0.38 & 0.77 & 0.61 \\
    DROID~\cite{droid} & Franka & End effector & 7 & 12.782 & 0.655 & 3.23 & 6.47 & 5.18 \\
    Bridge~\cite{bridgedata} & WidowX & End effector & 7 & 0.727 & 0.039 & 0.38 & 0.75 & 0.60 \\
    \midrule
    EgoVerse~\cite{egoverse} & Video only & --- & --- & 251.976 & 14.384 & 39.29 & 0 & 15.71 \\
    EgoDex~\cite{egodex} & Video only & --- & --- & 44.555 & 0.476 & 10.71 & 0 & 4.29 \\
    \midrule
    Total & & & & 557.928 & 20.195 & 100 & 100 & 100 \\
    \bottomrule
  \end{tabular*}
\end{table*}

\paragraph{Robot trajectories.}
We split by episode before constructing windows. We require finite state and gripper values, decodable camera streams, and alignment across views and between video and state. We remove stationary frames and motion outliers identified from arm velocity, acceleration, and jerk, discarding an episode if either fraction exceeds 95\%. Coarse motion-outlier screening thresholds are estimated per mapping and native frame rate across episodes passing initial integrity checks, whereas action and state normalization statistics are fitted strictly on training windows (using the 1st and 99th percentiles for arms and the minimum and maximum for grippers). Retained frames are compacted within each continuous segment, so windows never cross an episode boundary or a gap in the original recording.

\paragraph{Temporal and camera sampling.}
\begin{table*}[t]
  \centering
  \caption{Embodied pretraining hyperparameters by stage. Each optimizer step uses a global batch of 1,024, with one microbatch of 32 samples on each of 32 GPUs. Stage 1 learning rates are for the Video Expert.}
  \label{tab:supp-pt-opt}
  \small
  \setlength{\tabcolsep}{3pt}
  \begin{tabular*}{\textwidth}{@{\extracolsep{\fill}}cccc@{}}
    \toprule
    Configuration & Stage 1 (video only) & Stage 2 (action only) & Stage 3 (joint) \\
    \midrule
    Action and state bins & --- & 2,048 each & 2,048 each \\
    Robot / video-only stream probability (\%) & 50 / 50 & 100 / 0 & 80 / 20 \\
    \midrule
    Training steps & 80,000 & 40,000 & 200,000 \\
    Peak learning rate & $10^{-4}$ & $10^{-4}$ & $5\times10^{-5}$ \\
    Final learning rate & $10^{-5}$ & $5\times10^{-5}$ & $10^{-5}$ \\
    Schedule warmup (steps) & 1,000 & 1,000 & 2,000 \\
    \bottomrule
  \end{tabular*}
\end{table*}
\begin{figure*}[t]
  \centering
  \includegraphics[width=\textwidth]{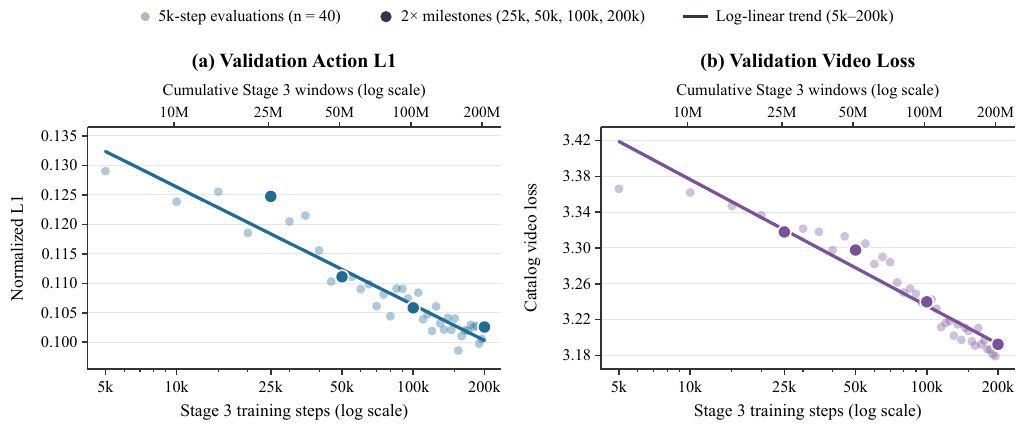}
  \caption{\textbf{Validation curves during Stage~3 joint pretraining.} All 40 positive-step evaluations (5k--200k steps, corresponding to 5.1M--204.8M seen windows at global batch size 1,024) are plotted on a logarithmic training-step axis with linear metric axes. Faint circles show individual 5k-step checkpoints; highlighted markers indicate successive $2\times$ training-budget milestones (25k, 50k, 100k, and 200k); solid lines show log-linear trend lines.}
  \label{fig:supp-stage3}
\end{figure*}
For InternData-A1, AgiBot World Alpha, EgoDex, and EgoVerse, each training episode is assigned one of four effective frame rates (30, 15, 10, or 5\,Hz) in proportions of 40, 30, 20, and 10\%, respectively. The corresponding strides on the retained sequence are 1, 2, 3, and 6. Short segments therefore contribute fewer windows at larger strides. RoboMIND uses 30\,Hz, DROID 15\,Hz, and Bridge 5\,Hz. Across pretraining stages, the head view measures $256\times320$ pixels and each auxiliary view measures $128\times160$ (height $\times$ width). Camera assignments follow the source hardware. For example, DROID places a second exterior view in the right auxiliary slot, while the video-only stream uses a head view and black-image tokens in both auxiliary slots.

\paragraph{Video-only data.}
EgoDex uses parts 1--5 for training and the official test split for validation. We hold out EgoVerse episodes by task and embodiment. Windows in this stream supervise future image tokens only.

\subsection{Sampling, optimization, and transfer}
We first draw a stream and then a mapping within that stream. For mapping $i$ in stream $s$, with $N_i$ training windows, the probability of drawing it in stage $k$ is
\begin{equation}
  P_k(i)=P_k(s)\,\frac{N_i^{0.75}}{\sum_{j\in s}N_j^{0.75}}.
\end{equation}
Table~\ref{tab:supp-pt-inventory} gives the resulting probabilities for each mapping.

\paragraph{Shared optimization settings.}
All pretraining and SFT runs in Tables~\ref{tab:supp-pt-opt} and~\ref{tab:supp-sft-opt} use AdamW with $(\beta_1,\beta_2)=(0.9,0.95)$, bfloat16 training precision, and cosine learning-rate decay after warmup. Weight decay is 0.01 for pretraining, RoboTwin, and real-world ALOHA SFT, and 0.05 for LIBERO SFT. We use cross-entropy (CE) for each supervised video or action token prediction.

\paragraph{Three-stage curriculum.}
\begin{table*}[t]
  \centering
  \caption{Supervised fine-tuning hyperparameters by benchmark. Step counts give the cosine decay horizon; online RL starts from the 10k-step checkpoint trained with VCQs on RoboTwin 8 tasks. The RoboTwin 50 and Real World (ALOHA) columns report transfer from Stage~3 (200k) embodied pretraining.}
  \label{tab:supp-sft-opt}
  \small
  \setlength{\tabcolsep}{3pt}
  \begin{tabular*}{\textwidth}{@{\extracolsep{\fill}}cccccc@{}}
    \toprule
    & & & \multicolumn{2}{c}{RoboTwin 50 tasks} & \\
    \cmidrule(lr){4-5}
    Configuration & LIBERO & RoboTwin 8 tasks & Full & C2R & Real World \\
    \midrule
    Initial weights & URSA & URSA & Stage 3 (200k) & Stage 3 (200k) & Stage 3 (200k) \\
    Proprioceptive input & None & Joint positions & Joint positions & Joint positions & Joint positions \\
    Valid action dimensions & 7 & 14 & 14 & 14 & 14 \\
    \midrule
    Global batch size & 128 & 1,024 & 1,024 & 256 & 256 \\
    Training / decay steps & 40,000 & 40,000 & 40,000 & 40,000 & 4,000 \\
    Peak learning rate & $3\times10^{-5}$ & $10^{-4}$ & $10^{-4}$ & $5\times10^{-5}$ & $5\times10^{-5}$ \\
    Final learning rate & $3\times10^{-6}$ & $10^{-5}$ & $10^{-5}$ & $10^{-5}$ & $10^{-5}$ \\
    Warmup steps & 200 & 1,000 & 1,000 & 500 & 500 \\
    Inference denoising steps & 4 & 4 & 16 & 16 & 4 \\
    \bottomrule
  \end{tabular*}
\end{table*}
Table~\ref{tab:supp-pt-opt} lists the settings for each stage. Stage 1 adapts the URSA-1.7B Video Expert~\cite{ursa}. Stage 2 freezes the Video Expert, initializes the Action transformer from URSA-0.6B, and restarts the robot data stream. For the first 5,000 steps, we train only the new state and action interfaces and prompts. We then unfreeze the Action transformer and bring its learning rate up to the shared schedule over the next 1,000 steps. Stage 3 loads both experts from the end of Stage 2 and restarts the optimizer and learning-rate schedule. Sampling resumes from the Stage 2 position in the robot stream and the Stage 1 position in the video-only stream. On robot windows, the joint objective gives equal weight to video and action losses and masks invalid action coordinates.

\paragraph{Supervised adaptation.}
Table~\ref{tab:supp-sft-opt} compares the shared LIBERO policy, the policy used to initialize RL on RoboTwin 8 tasks, the two transfer runs on RoboTwin 50 tasks, and the real-world ALOHA policy.

RoboTwin and real-world ALOHA use per-dimension min--max normalization. On RoboTwin 50 tasks (Full and C2R) and real-world ALOHA, fine-tuning independently inherits the Split Aloha input embeddings and output heads from Stage~3 (200k) joint pretraining and restarts the optimizer.

\subsection{Validation during joint pretraining}
We evaluate Stage~3 before its first update (step~0) and every 5k steps through 200k steps (41 checkpoints in total). Normalized action L1 uses a fixed selection of 1,485 episodes across 495 tasks and 13 mappings, with 16 inference steps, a greedy final step, and hierarchical macro-averaging. Video loss measures the robot catalog objective. Figure~\ref{fig:supp-stage3} plots all 40 positive-step evaluations against Stage~3 optimizer steps and cumulative seen windows (1,024 windows per step). From step~0 to 200k steps, normalized action L1 decreases by 16.9\% (from 0.1234 to 0.1026, minimum 0.0986 at 155k) and catalog video loss decreases by 3.9\% (from 3.321 to 3.192, minimum 3.179 at 195k), showing overall improvement in both objectives throughout joint pretraining.

%% file: sec/supp_rl.tex
\section{Reinforcement Learning Formulation and Implementation}
\label{sec:supp-rl}

\subsection{Recorded paths and likelihood replay}
Equation~\eqref{eq:aug_trace} defines the joint probability of a recorded
trace $\zeta=(x^{0:J},\hat x^{0:J-1})$, containing both noisy action states
and sampled clean proposals. With the conditioning input, noise prior,
and scheduler fixed, the prior and transition kernels are identical
under the current and behavior policies and cancel in the path
importance ratio. Importance weighting also requires the new policy's
support to be covered by the behavior policy; categorical proposals
have common support in exact arithmetic. This is a ratio for the
recorded augmented path. Evaluating the marginal probability of a final
action chunk would instead require summing over all paths ending in that
chunk.

Because every transition kernel $q_j$ in the fixed $J$-step URSA
schedule---including the final deterministic endpoint readout
$x^J=\hat x^{J-1}$ (Appendix~\ref{sec:supp-model})---is independent of
$\theta$, exact kernel cancellation in Eq.~\eqref{eq:aug_trace} holds
across the entire $J$-step trace, and the local PPO objective uniformly
scores all $J$ proposals along the recorded trace.

\begin{table}[t]
    \centering
    \small
    \caption{Core RL hyperparameters for the four RoboTwin
    eight-task configurations. Rollout batch size counts episodes;
    global minibatch size counts stored chunk slots.}
    \label{tab:supp-rl-settings}
    \setlength{\tabcolsep}{3pt}
    \begin{tabular}{@{}cc@{}}
        \toprule
        Setting & Value \\
        \midrule
        Action horizon / denoising steps & 16 / 4 \\
        Control-action budget per episode & 256 \\
        Rollout batch size (episodes) & 512 \\
        Global minibatch (chunk slots) & 512 \\
        PPO epochs per rollout batch & 1 \\
        GAE discount / trace parameter & 0.99 / 0.95 \\
        Actor / value-head learning rate & $2\times10^{-5}$ / $10^{-4}$ \\
        Global gradient norm limit & 1.0 \\
        PPO ratio clipping ($1\pm\epsilon$) & $[0.8,1.2]$ \\
        Log-ratio clipping ($\pm\delta_{\max}$) & $[-20,20]$ \\
        Number of VCQs & 80 \\
        \bottomrule
    \end{tabular}
\end{table}

\begin{table*}[t]
    \centering
    \small
    \caption{Reward and expert-update configurations for the four
    RoboTwin eight-task RL runs. All configurations retain VCQs;
    sparse-reward runs disable the dense reward term, giving an
    effective dense-reward weight of zero.}
    \label{tab:supp-rl-variants}
    \setlength{\tabcolsep}{8pt}
    \begin{tabular}{@{}ccccc@{}}
        \toprule
        Setting
        & \shortstack{Action-only\\Sparse}
        & \shortstack{Action-only\\Mixed}
        & \shortstack{Joint\\Sparse}
        & \shortstack{Joint\\Mixed} \\
        \midrule
        Video backbone & Frozen & Frozen & Trainable & Trainable \\
        Effective dense-reward weight ($\lambda_{\text{VCQ}}$) & 0 & 0.05 & 0 & 0.05 \\
        VCQ reward-score clipping ($[-g_{\max},g_{\max}]$) & --- & $[-1,1]$ & --- & $[-1,1]$ \\
        \bottomrule
    \end{tabular}
\end{table*}

\subsection{VCQ reward}
Following InternW0-$\Delta$~\cite{miao2026internw0delta}, the RoboTwin
eight-task SFT checkpoint includes 80 learnable VCQs. Each query sits at the center of a $2\times2$ patch of the
$16\times20$ head-view token grid. Let $z_t$ be the six bounded tokenizer
coordinates per location, divided componentwise by the FSQ half-levels
before rounding. Each query projects to $2\times2\times6$ values;
assembling the patches yields the predicted visual change
$\hat{\Delta z}_k$ from the current input. During SFT, the two clean-token
cross-entropies have unit weights, and an auxiliary MSE with weight 0.5
supervises the queries against the detached target $z_{t+16}-z_t$.

During rollout, the policy predicts visual change before drawing an
action and computes the observed latent change $\Delta z_k$ after
execution. The mixed reward follows Eq.~\eqref{eq:vcq_reward} with
stability constant $\epsilon_0 = 10^{-6}$ (and $\lambda_{\text{VCQ}}$,
$g_{\max}$ listed in Table~\ref{tab:supp-rl-variants}); its dense
term measures predictive gain relative to a zero-change baseline.
The task reward is one at first success and zero otherwise.
The mask $m_k^{\text{VCQ}}$ retains only active chunks ending before success
or the time limit. Terminal chunks still contribute their task reward
and actor loss.

\subsection{Training protocol and hyperparameters}
The four RoboTwin eight-task configurations combine action-only or joint
expert updates with sparse or mixed rewards. All start from the same
10k-step VCQ-equipped SFT checkpoint. Tables~\ref{tab:supp-rl-settings}
and~\ref{tab:supp-rl-variants} list the shared training settings and
reward and expert-update configurations, respectively.
We use AdamW without an entropy bonus or reference-policy KL penalty.

Each rollout batch contains 512 episodes, with 64 episodes per task.
We perform one PPO epoch per rollout batch, using a global minibatch
of 512 stored chunk slots.

\begin{enumerate}
\item \textit{Collect.} At each chunk, cache the context, compute the old
value, and sample all four clean proposals stochastically along a
denoising trace initialized from uniform valid action bins. Save the
conditioning input, noisy states, clean proposals, and old per-token
scores.
\item \textit{Execute.} Decode the final chunk and execute up to 16
actions. Record the next observation, first success, and truncation.
Compute the reward for the chosen variant.
\item \textit{Assign credit.} Compute chunk-level generalized advantages
with zero bootstrap at success and at the 256-action time limit.
Standardize advantages across valid samples independently on each rank.
\item \textit{Replay and update.} Shuffle chunk slots within each rank.
Rebuild condition features with gradients enabled, score the recorded
proposals at the recorded noisy states, and optimize the PPO actor
and value objectives. Rollout and replay use evaluation mode.
Synchronize weights to the rollout workers after each epoch over the batch.
\end{enumerate}

LIBERO uses the same proposal-path factorization and local clipping;
its suite-specific training and evaluation protocols are given in
Appendix~\ref{sec:supp-simulation}.

%% file: sec/supp_simulation.tex
\section{Simulation Benchmarks and Evaluation Protocols}
\label{sec:supp-simulation}

We compute SR by first averaging binary episode outcomes within
each task and then giving all tasks equal weight.

\paragraph{LIBERO.}
A single SFT policy covers all 40 tasks in the Spatial, Object, Goal and Long
suites~\cite{libero}, evaluated over 50 episodes per task (2,000 episodes in
total). Each suite-specific RL run starts from the same SFT checkpoint and
optimizes the policy under sparse task reward, collecting 512 online rollout
trajectories per update across the 10 tasks in the suite and reporting the peak suite-level online rollout success
rate. The mean across suites therefore summarizes one shared SFT policy or four
separately adapted RL policies. For context, dVLA-RL reports its best online rate separately for each LIBERO
task~\cite{dvlarl}.

\paragraph{RoboTwin 50 tasks.}
Both Full and C2R train one policy across all 50
tasks~\cite{robotwin}. Full uses 50 clean and 500 randomized demonstrations
per task (27,500 total), whereas C2R uses the 50 clean demonstrations per task
(2,500 total). Each policy is evaluated separately on clean and randomized
scenes over 100 episodes per task (5,000 episodes per condition), and Full and C2R are independent adaptations of their respective starting
weights.

\paragraph{RoboTwin 8 tasks.}
The eight tasks follow the manipulation subset of dVLA-RL~\cite{dvlarl} and
are listed in Table~\ref{tab:supp-rt8-tasks}. One VCQ-equipped SFT policy
learns from 1,000 randomized demonstrations per task (8,000 total). All four RL
configurations start from its 10,000-step checkpoint and use randomized scenes.
At SFT evaluation (\(R_0\)), action generation uses four denoising steps with
a greedy final proposal. The episode budget is 400 control actions for
six tasks, 300 for Handover Mic and 500 for Place Empty Cup. In online RL,
all four proposals are sampled stochastically and the budget is 256 control
actions. The policy predicts chunks of 16 actions and replans at chunk
boundaries, until success or timeout.

Each online rollout batch is collected prior to the corresponding PPO
update, and we report per-task scores from the batch achieving the largest
eight-task mean (Table~\ref{tab:supp-rt8-tasks}).

%% file: sec/supp_real_world.tex
\section{Real-World Evaluation on ALOHA}
\label{sec:supp-real-world}

\begin{figure}[t]
  \centering
  \includegraphics[trim=22bp 48bp 43bp 42bp,clip,width=0.72\linewidth]{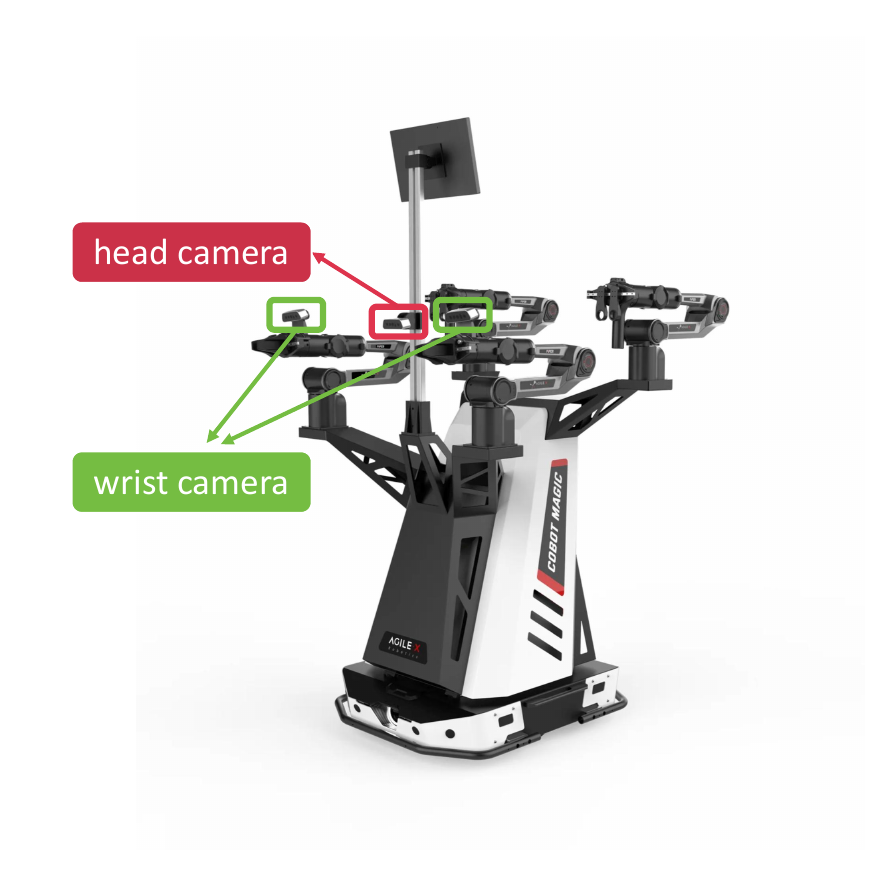}
  \caption{\textbf{Camera placement on the dual-arm ALOHA platform.} The red box marks the head camera and the green boxes mark the two wrist cameras.}
  \label{fig:supp-aloha-platform}
\end{figure}

\begin{figure*}[t]
  \centering
  \includegraphics[width=\textwidth]{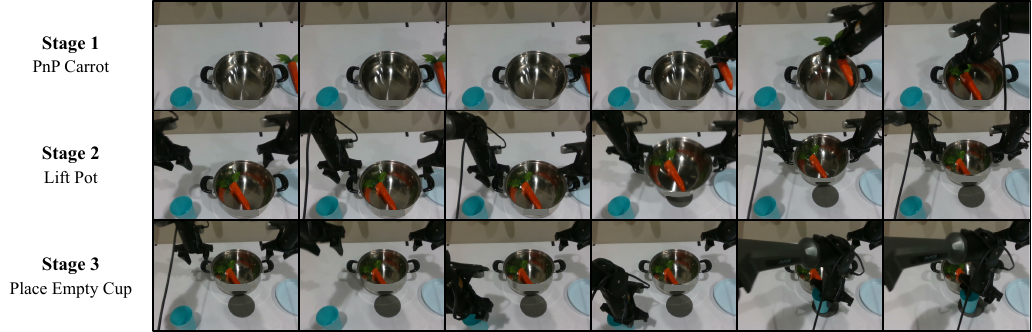}
  \caption{\textbf{Three sequential stages of \emph{Organize Table} on ALOHA.} Top to bottom: transferring the carrot into the pot (Stage~1), moving the pot with both arms to expose the coaster (Stage~2), and placing the empty cup on the coaster (Stage~3). Frames progress from left to right within each row.}
  \label{fig:supp-aloha-demo}
\end{figure*}

In \emph{Organize Table}, the dual-arm ALOHA robot~\cite{zhao2023learningfinegrainedbimanualmanipulation} (Figure~\ref{fig:supp-aloha-platform}) first transfers a carrot from a plate into a pot. It then grasps and moves the pot with both arms to uncover a coaster. The final stage requires placing an empty cup on the exposed coaster (Figure~\ref{fig:supp-aloha-demo}). Both DFM-WAM and $\pi_{0.5}$~\cite{pi05} were fine-tuned on the same 150 demonstrations. Each policy was evaluated in 50 trials of the complete three-stage task.

\paragraph{Training and deployment setup.}
DFM-WAM is initialized from the embodied-pretrained Stage~3 (200k) checkpoint, retaining the Split Aloha input embeddings and output heads while restarting the optimizer. Fine-tuning on the 150 \emph{Organize Table} demonstrations matches the RoboTwin C2R configuration (Table~\ref{tab:supp-sft-opt}): global batch size 256, 500 warmup steps, and cosine learning-rate decay from $5\times10^{-5}$ to $10^{-5}$ over 4,000 steps, with AdamW ($(\beta_1,\beta_2)=(0.9,0.95)$, weight decay 0.01), bfloat16 precision, cross-entropy targets, and per-dimension min--max normalization over the 14 dual-arm joint dimensions. At test time, the policy observes the head and two wrist cameras (Figure~\ref{fig:supp-aloha-platform}) and executes action chunks generated with 4-step DFM denoising using cached context and state K/V without generating future frames.

\paragraph{How stages are scored.}
When a stage fails, every subsequent stage in that trial is scored as a failure. The denominator is therefore 50 for every reported stage. Table~\ref{tab:supp-aloha} recovers the number of trials that passed each successive stage from the reported percentages and this denominator. Passing the cup stage requires having passed all three stages, so its rate is also the end-to-end task success rate.

\begin{table}[tb]
  \centering
  \small
  \caption{Trials passing each stage of \emph{Organize Table}. Each row counts trials that passed the indicated stage and all preceding stages.}
  \label{tab:supp-aloha}
  \begin{tabular}{@{}ccc@{}}
    \toprule
    Stages completed & $\pi_{0.5}$ & DFM-WAM \\
    \midrule
    Carrot transferred & 44/50 & 47/50 \\
    Pot moved, coaster exposed & 35/50 & 40/50 \\
    Cup placed, task complete & 34/50 & 37/50 \\
    \bottomrule
  \end{tabular}
\end{table}

DFM-WAM completed the full task in 37 of 50 trials (74\%), compared with 34 of 50 (68\%) for $\pi_{0.5}$. The largest difference in cumulative stage success was at the pot stage (80\% versus 70\%). The \emph{Avg.} reported in Table~\ref{tab:real_world_main} gives equal weight to the three cumulative stage rates, yielding 82.7\% and 75.3\%, respectively.

%% file: sec/supp_results.tex
\section{Additional Results and Ablations}
\label{sec:supp-results}

\begin{table*}[t]
\centering
\caption{Per-task success rates on RoboTwin 8 tasks (\%). The first four columns reproduce
Table~2 of dVLA-RL~\cite{dvlarl} with its published rounding.
OpenVLA-OFT and SimpleVLA-RL use single-task policies; the remaining
policies are multitask. DFM-WAM SFT ($R_0$) uses the SFT evaluation protocol
(100 randomized episodes per task, four denoising steps with greedy final proposal
and bin-center decoding). Each DFM-WAM online RL column ($R_1, R_2$) is taken
from the single online rollout batch that maximizes its eight-task mean (64
stochastic rollouts per task).}
\label{tab:supp-rt8-tasks}
\small
\setlength{\tabcolsep}{5pt}
\begin{tabular}{@{}cccccccc@{}}
\toprule
& \multicolumn{2}{c}{Reported SFT} & \multicolumn{2}{c}{Reported RL}
& \multicolumn{3}{c}{DFM-WAM (Ours)} \\
\cmidrule(lr){2-3}\cmidrule(lr){4-5}\cmidrule(l){6-8}
Task & $\pi_{0.5}$ & OpenVLA-OFT & SimpleVLA-RL & dVLA-RL & SFT ($R_0$) & Sparse ($R_1$) & Mixed ($R_2$) \\
\midrule
Beat Block Hammer & 88.0 & 28.1 & 87.5 & 95.3 & 74.00 & 98.44 & 95.31 \\
Place Phone Stand & 73.0 & 17.1 & 39.6 & 96.9 & 79.00 & 95.31 & 98.44 \\
Pick Dual Bottles & 98.0 & 29.7 & 68.3 & 95.3 & 90.00 & 90.62 & 95.31 \\
Lift Pot          & 88.0 & 10.1 & 64.1 & 89.1 & 96.00 & 98.44 & 95.31 \\
Move Can Pot      & 72.0 & 28.1 & 61.2 & 87.5 & 91.00 & 85.94 & 95.31 \\
Place A2B Left    & 74.0 & 37.5 & 45.3 & 79.7 & 73.00 & 95.31 & 90.62 \\
Place Empty Cup   & 74.0 & 77.3 & 94.2 & 95.3 & 92.00 & 93.75 & 95.31 \\
Handover Mic      & 32.0 & 45.3 & 89.2 & 96.9 & 71.00 & 85.94 & 84.38 \\
\midrule
Avg.              & 74.9 & 34.2 & 68.7 & 92.0 & 83.25 & 92.97 & 93.75 \\
\bottomrule
\end{tabular}
\end{table*}

\paragraph{Per-task results on RoboTwin 8 tasks.}
Table~\ref{tab:supp-rt8-tasks} reports per-task success rates for the
DFM-WAM SFT starting policy (\(R_0\)) and both RL variants (\(R_1, R_2\)).
Both RL variants substantially improve the average success rate over SFT
(\(R_0\): 83.25\%), outperforming \(R_0\) on seven of eight tasks each, with
mixed-reward RL (\(R_2\)) achieving the highest observed eight-task mean
(93.75\% versus 92.97\% for \(R_1\)).

\begin{figure*}[t]
    \centering
    \includegraphics[width=0.80\textwidth]
        {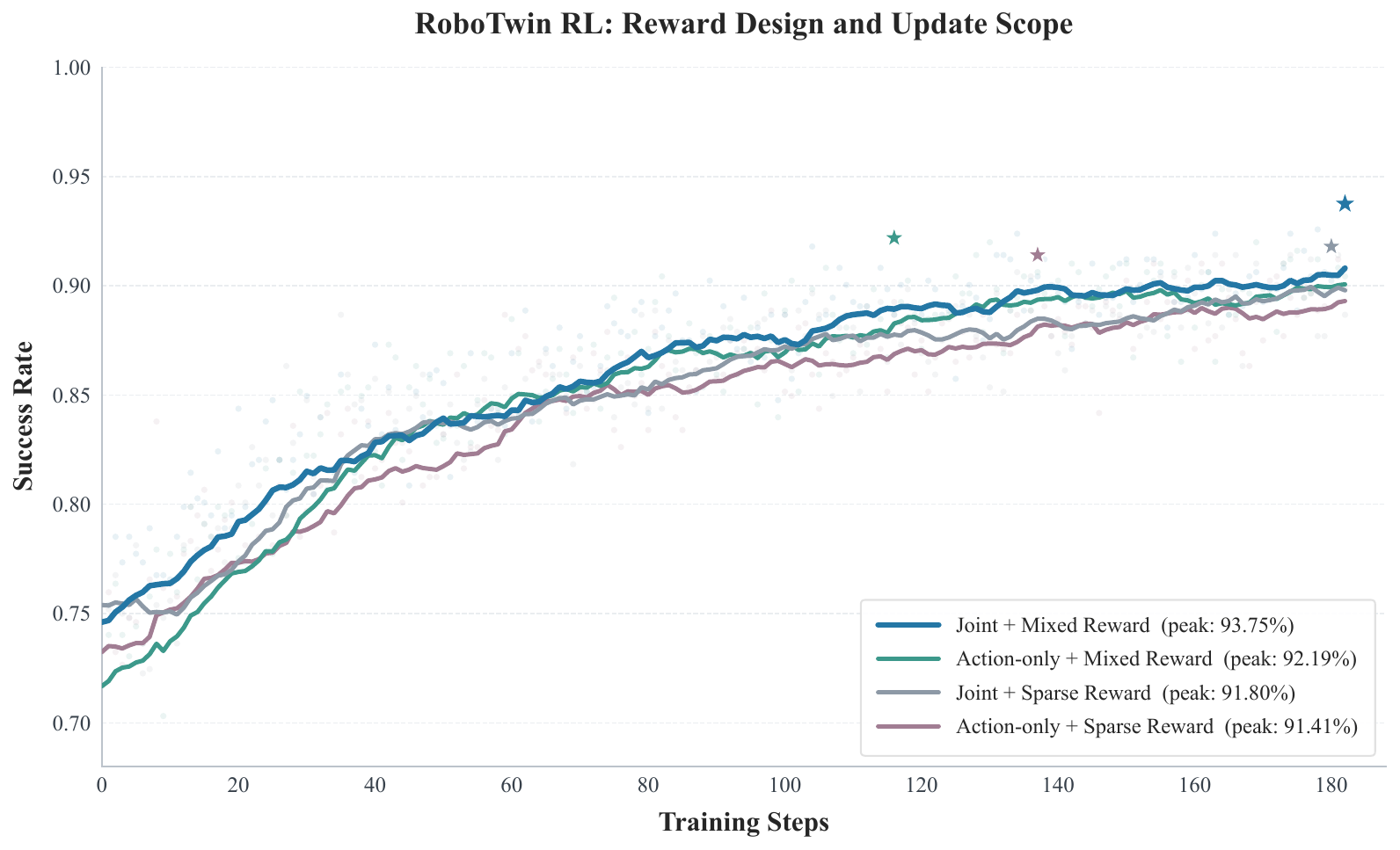}
    \caption{
    \textbf{Reward design and expert-update scope on RoboTwin eight tasks.}
    Each configuration is represented by a single training run.
    Points show raw online rollout success rates; solid curves show
    exponential moving averages with $\alpha=0.10$, using identical smoothing
    for all four configurations.
    All curves end at the joint mixed-reward run's observed peak.
    Stars mark the highest raw success rates within this shared window,
    with values reported in the legend.
    }
    \label{fig:supp-reward-scope}
\end{figure*}

\paragraph{Reward design and update scope.}
Figure~\ref{fig:supp-reward-scope} crosses sparse or mixed rewards with
action-only or joint expert updates on RoboTwin eight tasks
(Appendix~\ref{sec:supp-rl}). Tables~\ref{tab:supp-rl-settings}
and~\ref{tab:supp-rl-variants} report the shared training settings and
reward and expert-update configurations.
For a common comparison window, all curves are truncated at the observed
peak of the joint mixed-reward run, the 183rd rollout batch.
Within this window, mixed rewards yield higher observed peak rollout
success under both update scopes: 92.19\% versus 91.41\% for action-only
updates, and 93.75\% versus 91.80\% for joint updates, corresponding to
gains of 0.78 and 1.95 percentage points, respectively.
Continuing beyond this shared window, the action-only sparse run ($R_1$)
reaches its full-run peak of 92.97\% at rollout batch~215
(Table~\ref{tab:supp-rt8-tasks}).

\paragraph{Transfer across scene conditions.}
Under the Full protocol on RoboTwin 50 tasks, embodied-pretrained adaptation improves SR by 3.46
points on clean scenes and 2.76 on randomized scenes. After clean-only
SFT, the corresponding differences are 20.16 and 29.12 points. Even with
embodied pretraining, C2R success falls from 77.84\% on clean scenes to
31.64\% under randomization.

Table~\ref{tab:supp-robotwin50-full} gives the task-level breakdown for the
Full setting. For LingBot-VA, the integer per-task scores reproduced from
Fast-WAM~\cite{fastwam} (v2, Table~3) average to 92.90\% (clean) and
91.50\% (randomized), whereas Table~\ref{tab:exp-sft} quotes 92.93\% and
91.55\% from the original LingBot-VA report~\cite{lingbotva}.

\begin{table*}[p]
\centering
\caption{Success rates on RoboTwin 50 tasks under the Full protocol (\%). Full training uses 50 clean
and 500 randomized demonstrations per task. The task order and four
baseline results follow Fast-WAM~\cite{fastwam} (v2, Table~3).
The average weights all 50 tasks equally.}
\label{tab:supp-robotwin50-full}
\small
\setlength{\tabcolsep}{3pt}
\begin{tabular*}{\textwidth}{@{\extracolsep{\fill}}*{11}{c}@{}}
\toprule
 & \multicolumn{2}{c}{DFM-WAM (Ours)} & \multicolumn{2}{c}{Fast-WAM}
 & \multicolumn{2}{c}{$\pi_{0.5}$} & \multicolumn{2}{c}{LingBot-VA}
 & \multicolumn{2}{c}{Motus} \\
\cmidrule(lr){2-3}\cmidrule(lr){4-5}\cmidrule(lr){6-7}\cmidrule(lr){8-9}\cmidrule(l){10-11}
Task & Clean & Rand. & Clean & Rand. & Clean & Rand. & Clean & Rand. & Clean & Rand. \\
\midrule
Adjust Bottle & 100 & 98 & 100 & 100 & 100 & 99 & 90 & 94 & 89 & 93 \\
Beat Block Hammer & 80 & 88 & 99 & 97 & 96 & 93 & 96 & 98 & 95 & 88 \\
Blocks Ranking RGB & 98 & 99 & 100 & 100 & 92 & 85 & 99 & 98 & 99 & 97 \\
Blocks Ranking Size & 89 & 82 & 94 & 98 & 49 & 26 & 94 & 96 & 75 & 63 \\
Click Alarmclock & 99 & 100 & 100 & 100 & 98 & 89 & 99 & 100 & 100 & 100 \\
Click Bell & 100 & 100 & 100 & 100 & 99 & 66 & 100 & 100 & 100 & 100 \\
Dump Bin Bigbin & 95 & 100 & 97 & 96 & 92 & 97 & 89 & 96 & 95 & 91 \\
Grab Roller & 100 & 100 & 100 & 100 & 100 & 100 & 100 & 100 & 100 & 100 \\
Handover Block & 86 & 80 & 95 & 81 & 66 & 57 & 99 & 78 & 86 & 73 \\
Handover Mic & 93 & 95 & 99 & 100 & 98 & 97 & 94 & 96 & 78 & 63 \\
Hanging Mug & 70 & 63 & 58 & 62 & 18 & 17 & 40 & 28 & 38 & 38 \\
Lift Pot & 100 & 99 & 100 & 100 & 96 & 85 & 100 & 99 & 96 & 99 \\
Move Can Pot & 95 & 99 & 90 & 88 & 51 & 55 & 94 & 97 & 34 & 74 \\
Move Pillbottle Pad & 100 & 98 & 100 & 99 & 84 & 61 & 99 & 99 & 93 & 96 \\
Move Playingcard Away & 100 & 99 & 100 & 100 & 96 & 84 & 100 & 99 & 100 & 96 \\
Move Stapler Pad & 72 & 68 & 77 & 64 & 56 & 42 & 91 & 79 & 83 & 85 \\
Open Laptop & 95 & 100 & 98 & 100 & 90 & 96 & 92 & 94 & 95 & 91 \\
Open Microwave & 95 & 100 & 62 & 45 & 34 & 77 & 82 & 86 & 95 & 91 \\
Pick Diverse Bottles & 76 & 79 & 80 & 85 & 81 & 71 & 89 & 82 & 90 & 91 \\
Pick Dual Bottles & 83 & 86 & 100 & 96 & 93 & 63 & 100 & 99 & 96 & 90 \\
Place A2B Left & 99 & 98 & 95 & 93 & 87 & 82 & 97 & 93 & 88 & 79 \\
Place A2B Right & 98 & 99 & 93 & 99 & 87 & 84 & 97 & 95 & 91 & 87 \\
Place Bread Basket & 96 & 94 & 91 & 93 & 77 & 64 & 97 & 95 & 91 & 94 \\
Place Bread Skillet & 96 & 94 & 90 & 93 & 85 & 66 & 95 & 90 & 86 & 83 \\
Place Burger Fries & 98 & 98 & 96 & 99 & 94 & 87 & 97 & 95 & 98 & 98 \\
Place Can Basket & 81 & 74 & 71 & 69 & 62 & 62 & 81 & 84 & 81 & 76 \\
Place Cans Plasticbox & 96 & 97 & 99 & 96 & 94 & 84 & 100 & 99 & 98 & 94 \\
Place Container Plate & 99 & 100 & 96 & 100 & 99 & 95 & 99 & 97 & 98 & 99 \\
Place Dual Shoes & 88 & 78 & 94 & 88 & 75 & 75 & 94 & 89 & 93 & 87 \\
Place Empty Cup & 100 & 100 & 100 & 100 & 100 & 99 & 100 & 100 & 99 & 98 \\
Place Fan & 95 & 91 & 96 & 96 & 87 & 85 & 99 & 93 & 91 & 87 \\
Place Mouse Pad & 94 & 94 & 83 & 89 & 60 & 39 & 93 & 96 & 66 & 68 \\
Place Object Basket & 86 & 83 & 89 & 88 & 80 & 76 & 91 & 88 & 81 & 87 \\
Place Object Scale & 96 & 96 & 90 & 97 & 86 & 80 & 96 & 95 & 88 & 85 \\
Place Object Stand & 100 & 95 & 90 & 94 & 91 & 85 & 99 & 96 & 98 & 97 \\
Place Phone Stand & 64 & 68 & 97 & 99 & 81 & 81 & 97 & 97 & 87 & 86 \\
Place Shoe & 97 & 99 & 96 & 99 & 92 & 93 & 98 & 98 & 99 & 97 \\
Press Stapler & 99 & 100 & 90 & 97 & 87 & 83 & 85 & 82 & 93 & 98 \\
Put Bottles Dustbin & 76 & 71 & 95 & 90 & 84 & 79 & 87 & 91 & 81 & 79 \\
Put Object Cabinet & 90 & 85 & 94 & 89 & 80 & 79 & 85 & 87 & 88 & 71 \\
Rotate QRcode & 88 & 88 & 93 & 89 & 89 & 87 & 96 & 91 & 89 & 73 \\
Scan Object & 89 & 88 & 89 & 92 & 72 & 65 & 96 & 91 & 67 & 66 \\
Shake Bottle & 99 & 98 & 100 & 100 & 99 & 97 & 100 & 97 & 100 & 97 \\
Shake Bottle Horizontally & 99 & 99 & 100 & 100 & 99 & 99 & 100 & 99 & 100 & 98 \\
Stack Blocks Three & 98 & 90 & 95 & 97 & 91 & 76 & 99 & 98 & 91 & 95 \\
Stack Blocks Two & 100 & 100 & 100 & 100 & 97 & 100 & 100 & 98 & 100 & 98 \\
Stack Bowls Three & 80 & 78 & 80 & 81 & 77 & 71 & 86 & 83 & 79 & 87 \\
Stack Bowls Two & 99 & 92 & 92 & 98 & 95 & 96 & 94 & 98 & 98 & 98 \\
Stamp Seal & 88 & 95 & 90 & 94 & 79 & 55 & 96 & 97 & 93 & 92 \\
Turn Switch & 82 & 76 & 61 & 59 & 62 & 54 & 44 & 45 & 84 & 78 \\
\midrule
Avg. & 91.92 & 91.02 & 91.88 & 91.78 & 82.74 & 76.76 & 92.90 & 91.50 & 88.66 & 87.02 \\
\bottomrule
\end{tabular*}
\end{table*}

\subsection{Expert initialization and future-frame supervision on RoboTwin 8 tasks}
\label{sec:supp-sft-ablations}
\begin{figure}[tb]
\centering
\includegraphics[width=\linewidth]{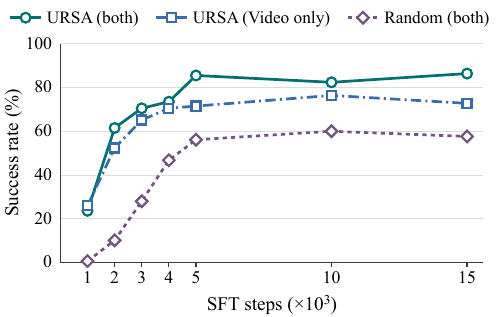}
\caption{\textbf{Expert initialization on RoboTwin 8 tasks.} All three runs jointly
supervise future frames and actions. Each point gives closed-loop success
over 800 episodes (100 per task).}
\label{fig:supp-rt8-sft-init}
\end{figure}
We evaluate four multitask SFT runs at seven checkpoints from 1k to 15k
steps over equal numbers of optimizer steps. All runs use CE and omit VCQs.
Each checkpoint is evaluated on 100 episodes per task in randomized scenes
with unseen instructions. At each observation, the policy caches the current
visual context and denoises a 16-step action chunk in four denoising steps,
taking the final proposal greedily. None of the policies generates future
frames during evaluation.

\paragraph{Expert initialization.}
Figure~\ref{fig:supp-rt8-sft-init} compares three runs trained with the same
joint video and action objective and pretrained visual tokenizer. With the
Action Expert initialized randomly, transferring the Video Expert from URSA
gives higher success at every measured checkpoint. At 5k steps the gap is
15.4 percentage points (71.5\% versus 56.1\%). With the Video Expert
initialized from URSA in both runs, transferring the Action Expert as well
yields 85.5\% at 5k (a further 14.0-point gain) and reaches 86.4\% at 15k.

\begin{figure}[tb]
\centering
\includegraphics[width=\linewidth]{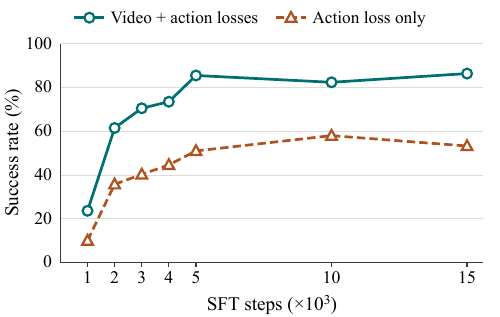}
\caption{\textbf{Future-frame co-training on RoboTwin 8 tasks with both experts initialized
from URSA.} The video + action curve is the same run as
\emph{URSA (both)} in Figure~\ref{fig:supp-rt8-sft-init}. Each point gives
closed-loop success over 800 episodes.}
\label{fig:supp-rt8-sft-cotrain}
\end{figure}

\paragraph{Future-frame co-training.}
Figure~\ref{fig:supp-rt8-sft-cotrain} holds both experts' URSA initialization
fixed and varies whether SFT also supervises the future frame. The Video
Expert receives gradients from the action loss in both runs. Adding the
future-frame loss improves success at every measured checkpoint. At 5k steps,
the jointly supervised policy succeeds in 85.5\% of episodes, compared with
51.0\% for action supervision alone, and scores higher on seven of eight
tasks. Handover Mic shows the largest difference (68 versus 8 successes in
100 episodes), while both policies nearly saturate Lift Pot (98 versus 99).
At 15k, joint supervision retains a 33.1-point advantage (86.4\% versus
53.3\%). The benefit of predicting future frames during training therefore
persists when the policy executes actions using only the current context.

\begin{figure*}[t]
\centering
\includegraphics[width=\textwidth]{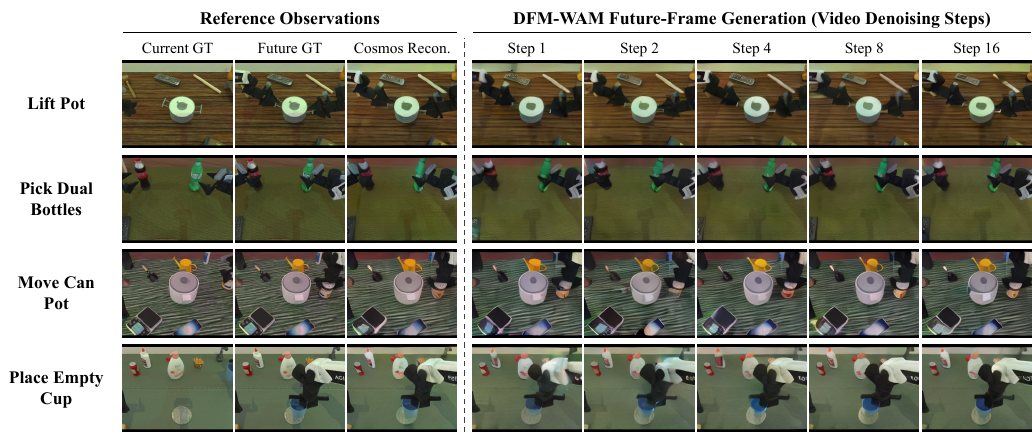}
\caption{\textbf{Qualitative future-frame generation on RoboTwin 50 tasks.} Predictions are generated by the embodied-pretrained Full model (transferred from Stage~3 200k). Left to right: current ground-truth head observation (\emph{Current GT}), future ground-truth observation 16 steps ahead (\emph{Future GT}), Cosmos DI16$\times$16 tokenizer reconstruction of \emph{Future GT} (\emph{Cosmos Recon.}), and DFM-WAM Video Expert generations with 1, 2, 4, 8, and 16 discrete flow denoising steps.}
\label{fig:supp-video-gen}
\end{figure*}

\subsection{Qualitative future-frame generation on RoboTwin 50 tasks}
\label{sec:supp-video-gen}
Although test-time control omits the future-frame branch and executes cached action-only inference (Appendix~\ref{sec:supp-model}), the Video Expert is trained to generate future head-camera frames 16 steps ahead along the metric discrete flow path. Figure~\ref{fig:supp-video-gen} visualizes future-frame predictions from the embodied-pretrained RoboTwin 50 tasks Full model (initialized from Stage~3 200k) on four representative tasks spanning bimanual and single-arm manipulation under randomized scenes: \emph{Lift Pot}, \emph{Pick Dual Bottles}, \emph{Move Can Pot}, and \emph{Place Empty Cup}.

For each task, we compare the current observation (\emph{Current GT}), the ground-truth frame 16 steps later (\emph{Future GT}), the frozen Cosmos DI16$\times$16 reconstruction of \emph{Future GT} (\emph{Cosmos Recon.}), and Video Expert generations under 1, 2, 4, 8, and 16 discrete flow denoising steps. Across these denoising budgets, the Video Expert captures the major scene layout, manipulator motion, and object transitions, with visual detail becoming cleaner as more denoising steps are used---even though future-frame generation is omitted during test-time control.

%% file: sec/supp_baselines.tex
\section{Baseline Variants and Sources}
\label{sec:supp-baselines}

\begin{table}[t]
  \centering
  \small
  \setlength{\tabcolsep}{2pt}
  \caption{Baseline variants and source tables for LIBERO and RoboTwin Full in Table~\ref{tab:exp-sft}. Entries cite the arXiv revision table used for each score (including secondary compilations~\cite{openvlaoft,fastwam,imagewam}). $^\dagger$MM-ACT uses Vanilla on Spatial, Object, and Goal, and +Text on Long~\cite{mmact}.}
  \label{tab:supp-sft-sources}
  \begin{tabular}{@{}ccc@{}}
    \toprule
    Selected policy (variant) & LIBERO & RoboTwin Full \\
    \midrule
    $\pi_0$ (fine-tuned) & v2, Tab.~I~\cite{openvlaoft} & v1, Tab.~1~\cite{imagewam} \\
    $\pi_{0.5}$ & v1, Tab.~2~\cite{imagewam} & v1, Tab.~1~\cite{imagewam} \\
    OpenVLA-OFT & v2, Tab.~I~\cite{openvlaoft} & --- \\
    $\pi_0$-FAST (fine-tuned) & v2, Tab.~I~\cite{openvlaoft} & --- \\
    OpenVLA & v3, Tab.~12~\cite{openvla} & --- \\
    DFM-VLA (+Embed) & v1, Tab.~2~\cite{dfmvla} & --- \\
    DD-VLA & v4, Tab.~1~\cite{ddvla} & --- \\
    MM-ACT$^\dagger$ & v2, Tab.~1~\cite{mmact} & --- \\
    \midrule
    Fast-WAM (default) & v1, Tab.~2~\cite{fastwam} & v1, Tab.~1~\cite{fastwam} \\
    Cosmos Policy (no planning) & v1, Tab.~1~\cite{cosmospolicy} & --- \\
    LingBot-VA (pretrained) & v2, Tab.~2~\cite{lingbotva} & v2, Tab.~1~\cite{lingbotva} \\
    Motus (pretrained) & v1, Tab.~2~\cite{fastwam} & v2, Tab.~14~\cite{motus} \\
    ImageWAM (FLUX.2 4B) & v1, Tab.~2~\cite{imagewam} & v1, Tab.~1,12~\cite{imagewam} \\
    WorldVLA ($512\times512$) & v1, Tab.~2~\cite{worldvla} & --- \\
    MMaDA-VLA & v3, Tab.~1~\cite{mmadavla} & --- \\
    \bottomrule
  \end{tabular}
\end{table}

\paragraph{Sources and training scope.}
Table~\ref{tab:supp-sft-sources} records the selected variant and source table for each external baseline in Table~\ref{tab:exp-sft}. Training scope varies across LIBERO sources: OpenVLA averages three seeds of suite-specific fine-tuning~\cite{openvla} and OpenVLA-OFT likewise trains suite-specific policies~\cite{openvlaoft}, whereas DFM-WAM uses a single shared policy across all four suites. On RoboTwin C2R, the three baseline rows ($\pi_0$, $\pi_{0.5}$, and Fast-WAM) come from the official RoboTwin~2.0 leaderboard~\cite{robotwin} (snapshot of 15 November 2026) and were trained for C2R separately from the published Full models: $\pi_0$ uses one policy per task, whereas $\pi_{0.5}$ and Fast-WAM each share one 50-task policy.

\paragraph{Parameter counts.}
Model sizes follow the convention of Efficient-WAM~\cite{efficientwam}. For DFM-WAM, 2.3B sums the nominal URSA checkpoints initializing the Video (1.7B) and Action (0.6B) Experts. Among compared WAMs, Cosmos Policy's 2B counts the shared Cosmos-Predict2 denoiser excluding the text encoder and VAE~\cite{cosmospolicy}; Fast-WAM counts 5B video and 1B action parameters~\cite{fastwam}; ImageWAM's 4.6B combines its 4B image branch and 642M action expert~\cite{imagewam}; and Motus includes its frozen VLM in 8B~\cite{motus}.